\pdfoutput=1

\documentclass[11pt]{article}
\usepackage{float}

\usepackage[a-1b]{pdfx}

\usepackage{framed}
\usepackage{mdwlist}
\usepackage{siunitx}
\usepackage{latexsym}
\usepackage{colortbl}
\usepackage{xcolor}
\usepackage{nicefrac}
\usepackage{booktabs}
\usepackage{fnpct}
\usepackage{amsfonts}
\usepackage[T1]{fontenc}
\usepackage{bold-extra}
\usepackage{amsmath}
\usepackage{amssymb}
\usepackage{bm}
\usepackage{graphicx}
\usepackage{mathtools}
\usepackage{microtype}
\usepackage{multirow}
\usepackage{multicol}
\usepackage{xpatch}
\usepackage{latexsym,comment}
\usepackage[normalem]{ulem}

\newcommand*{\missingreference}{{\Huge \colorbox{red}{?reference?}}}
\newcommand*{\missingcitation}{{\Huge \colorbox{red}{?citation?}}}

\makeatletter
\xpatchcmd{\@setref}{\bfseries}{\missingreference}{}{}
\def\@citex[#1]#2{\leavevmode
    \let\@citea\@empty
    \@cite{\@for\@citeb:=#2\do
        {\@citea\def\@citea{,\penalty\@m\ }%
            \edef\@citeb{\expandafter\@firstofone\@citeb\@empty}%
            \if@filesw\immediate\write\@auxout{\string\citation{\@citeb}}\fi
            \@ifundefined{b@\@citeb}{\hbox{\reset@font\missingcitation}%
                \G@refundefinedtrue
                \@latex@warning
                {Citation `\@citeb' on page \thepage \space undefined}}%
            {\@cite@ofmt{\csname b@\@citeb\endcsname}}}}{#1}}
\makeatother

\newcommand{\gem}[1]{\mbox{\textsc{gem}}}
\newcommand{\abr}[1]{\textsc{#1}}

\newcommand{\g}{\, | \,}

\newcommand{\hidetext}[1]{}
\newcommand{\ignore}[1]{}
\newif\ifcomment
\commenttrue

\ifcomment
    \newcommand{\pinaforecomment}[3]{\colorbox{#1}{\parbox{.8\linewidth}{#2: #3}}}

    \newcommand{\prtodo}[1]{\pinaforecomment{lightblue}{pr}{#1}}
    \newcommand{\prtodoi}[1]{\pinaforecomment{lightblue}{pr}{#1}}
\else
    \newcommand{\pinaforecomment}[3]{}
    \newcommand{\prtodo}[1]{}
    \newcommand{\prtodoi}[1]{}
\fi

\newcommand{\smallurl}[1]{ \begin{tiny}\url{#1}\end{tiny}}

\definecolor{lightblue}{HTML}{3cc7ea}
\definecolor{CUgold}{HTML}{CFB87C}
\definecolor{grey}{rgb}{0.95,0.95,0.95}
\definecolor{ceil}{rgb}{0.57, 0.63, 0.81}
\definecolor{UMDred}{HTML}{ed1c24}
\definecolor{UMDyellow}{HTML}{ffc20e}

\usepackage[]{acl2023}

\usepackage{xspace}
\usepackage[utf8]{inputenc}
\usepackage{pgfplots}
\usepackage{dsfont}
\DeclareUnicodeCharacter{2212}{−}
\usepgfplotslibrary{groupplots,dateplot}
\usetikzlibrary{patterns,shapes.arrows}
\pgfplotsset{compat=newest}

\usepackage{tikzscale}
\usepackage{amsmath,amssymb}
\newcommand{\probP}{\text{I\kern-0.15em P}}
\newcommand{\KARL}{$\text{KAR}^\text{3}\text{L}$\xspace}
\newcommand{\model}{\textsc{Smart}\xspace}
\newcommand{\mnemonic}{m}

\usepackage{multirow, colortbl}

\usepackage{tabularx,booktabs}
\usepackage{makecell}

\usepackage[normalem]{ulem}
\useunder{\uline}{\ul}{}

\usepackage{graphicx}

\definecolor{ablation6}{HTML}{fcefed}
\definecolor{ablation_tie}{HTML}{fce3e1}

\definecolor{ablation5}{HTML}{fcd8d4}
\definecolor{ablation4}{HTML}{FBC3BC}
\definecolor{ablation3}{HTML}{F7A399}
\definecolor{ablation2}{HTML}{F38375}
\definecolor{ablation1}{HTML}{EF6351}

\usepackage[]{algpseudocode}
\usepackage[]{algorithm}
\usepackage{float}
\algtext*{EndFor}%
\algtext*{EndProcedure}%

\usepackage{booktabs}
\usepackage[normalem]{ulem}
\useunder{\uline}{\ul}{}

\usepackage{times}
\usepackage{latexsym}
\usepackage{adjustbox}

\usepackage[T1]{fontenc}
\usepackage{amsmath}
\usepackage{amssymb}
\usepackage{booktabs}
\usepackage{tikzscale}
\usepackage{amsmath}

\newcommand{\specialcellleft}[2][l]{%
\begin{tabular}[#1]{@{}l@{}}#2\end{tabular}}

\usepackage{multirow, colortbl}

\usepackage{tabularx,booktabs}
\usepackage{makecell}
\usepackage{multirow}
\usepackage{scalerel,xparse}
\usepackage[utf8]{inputenc}

\usepackage{cleveref}
\usepackage{microtype}
\usepackage{enumitem}
\crefformat{section}{\S#2#1#3}
\crefformat{subsection}{\S#2#1#3}
\crefformat{subsubsection}{\S#2#1#3}

\definecolor{UMDred}{HTML}{ed1c24}

\title{A \model Mnemonic Sounds like \textit{``Glue Tonic''}:\\ Mixing LLMs with Student Feedback to Make Mnemonic Learning Stick}

\author{Nishant Balepur$^{1}$ \hspace{0.5cm} Matthew Shu$^{2}$  \hspace{0.5cm} \textbf{Alexander Hoyle}$^{1}$ \hspace{0.5cm} \textbf{Alison Robey}$^{3}$ \\ \hspace{0.5cm} \textbf{Shi Feng}$^{4}$ \hspace{0.5cm} \textbf{Seraphina Goldfarb-Tarrant}$^{5}$ \hspace{0.5cm} \textbf{Jordan Boyd-Graber}$^{1}$ \vspace{0.1cm}  \\
  $^{1}$University of Maryland \hspace{0.5cm}
  $^{2}$Yale University \hspace{0.5cm}
  $^{3}$SUNY Empire State University \hspace{0.5cm} \\
  $^{4}$George Washington University \hspace{0.5cm}
  $^{5}$Cohere \hspace{0.5cm} \vspace{0.1cm} \\
  \texttt{nbalepur@umd.edu} \hspace{0.5cm} \texttt{{jbg}@umiacs.umd.edu}
}

\begin{document}
\maketitle

\begin{abstract} {
Keyword mnemonics are memorable explanations that link new terms to simpler keywords.
Prior work generates mnemonics for students, but they do not train models using mnemonics students prefer and aid learning.
We build \model, a mnemonic generator trained on feedback from real students learning new terms.
To train \model, we first fine-tune LLaMA-2 on a~curated set of user-written mnemonics.
We then use LLM alignment to enhance \model: we deploy mnemonics generated by \model in a flashcard app to find preferences on mnemonics students favor.
We gather 2684 preferences from 45 students across two types: \textbf{expressed} (inferred from ratings) and \textbf{observed} (inferred from student learning), yielding three key findings.
First, expressed and observed preferences disagree; what students \textit{think} is helpful does not always capture what is \textit{truly} helpful.
Second, Bayesian models can synthesize complementary data from multiple preference types into a single effectiveness signal.
\model is tuned via Direct Preference Optimization on this signal, which resolves ties and missing labels in the typical method of pairwise comparisons, augmenting data for LLM output quality gains. 
Third, mnemonic experts assess \model as matching GPT-4 at much lower deployment costs, showing the utility of capturing diverse student feedback to align LLMs in education.\footnote{ \url{https://github.com/nbalepur/Mnemonic}}
}
\end{abstract}

\section{Mnemonics Aid Vocabulary Learning}
\begin{figure*}
    \centering
    \includegraphics[width=\linewidth]{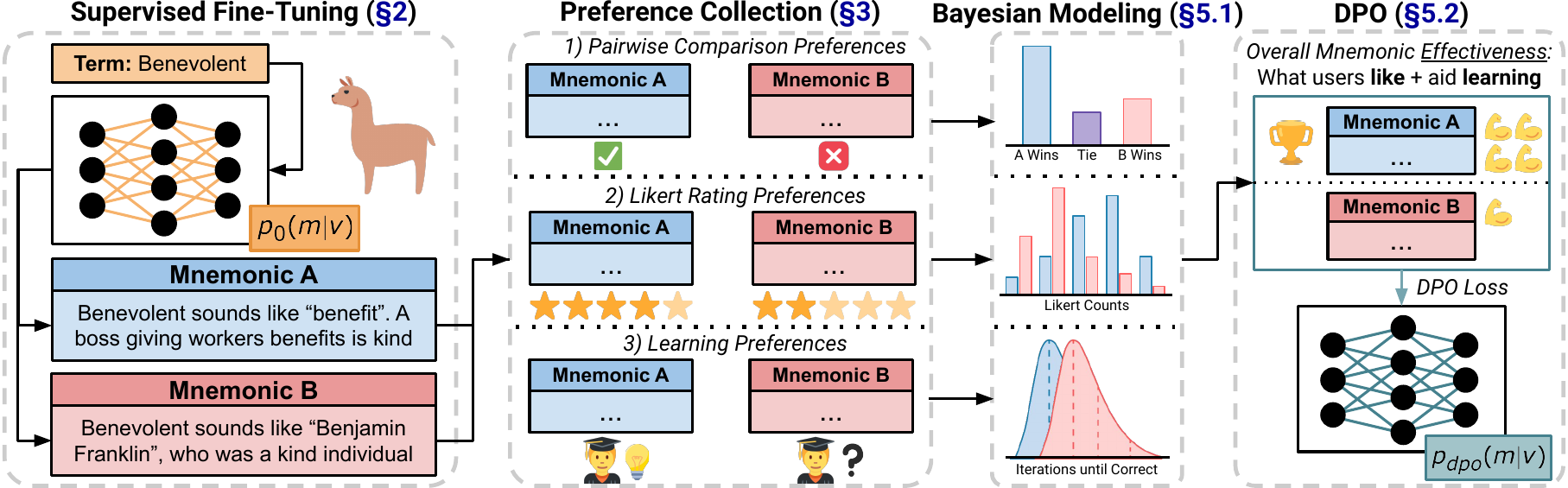}
    \caption{\model overview. We fine-tune LLaMA-2 70B for the initial \model model (\cref{section:initial_model}). We then~collect three preference types: pairwise, rating, and learning (\cref{section:user_study}). Finally, a Bayesian model synthesizes mnemonic~effectiveness from all three preferences (\cref{subsection:bayesian_rlhf}) and we use this signal to align \model via Direct Preference Optimization (\cref{subsection:DPO}).}
    \label{fig:intro}
\end{figure*}
Keyword mnemonics promote efficient and engaging vocabulary (vocab) learning \cite{benge2009using}.
These tools help students learn a new term's meaning (e.g. \textit{\textbf{Benevolent}}) by relating it to a simpler keyword (e.g. \textit{Benevolent sounds like~\textbf{benefit}}), and explaining how the keyword and term are linked (e.g. \textit{A boss giving employee \textbf{benefits} is kind, which is the meaning of \textbf{benevolent}}) \cite{pressley1982mnemonic}.
Students use mnemonics to prepare for exams like the GRE \cite{fairbanks1977vocabulary} which involve mastering hundreds of terms \cite{alstudied}.
Despite their utility, writing mnemonics is tedious, requiring vocabulary expertise and creativity to~make memorable keyword links \cite{siriganjanavong2013mnemonic}.

To ease these burdens, prior work automatically generates keyword mnemonics \cite{savva2014transphoner}.
However, most works design keyword extractors \cite{7409024}, omitting the explanations linking keywords to terms that enable effective mnemonic use \cite{raugh1975mnemonic}. 
Large Language Models (LLMs) are apt for writing explanations, a difficult task that tests if LLMs can combine vocabulary \cite{huang-etal-2022-understanding}, phonology \cite{suvarna2024phonologybench}, commonsense \cite{davis2023benchmarks}, and creativity \cite{tian2023macgyver} to help students learn (\cref{subsection:qual}).
While promising, existing works only \textit{prompt} LLMs \cite{lee2023smartphone}~and~lack training on \textit{student feedback} to guide LLMs toward mnemonics students prefer and benefit learning.

In pursuit of student-guided mnemonics, we propose \model, which employs \textbf{S}tudent \textbf{M}nemonic \textbf{A}lignment to generate keyword mnemonics that aid the \textbf{R}ecall of \textbf{T}erms. (Figure~\ref{fig:intro}). 
To train \model, we first get data from MnemonicDictionary \cite{mnemonic_dictionary}, a site where users submit mnemonics that they find helpful.
We collect a high-quality subset of submitted mnemonics to fine-tune LLaMA-2~70B (Figure~\ref{fig:intro}, left) as our initial model (\cref{section:initial_model}).
To enhance \model, we draw from LLM alignment, which improves LLMs via tuning to preference labels that capture which of two LLM outputs users favor \cite{casper2023open}.
We gather preferences by sampling mnemonics from our initial model and deploying them to students in a flashcard app (\cref{section:user_study}).


There are many ways to collect preferences, and \citet{bansal2023peering} show that pairwise rankings and Likert ratings yield conflicting labels on which LLM output is favored.
To study preference agreement in education (Figure~\ref{fig:intro}, mid), we also gather pairwise and Likert annotations, which we define as \textbf{expressed preferences:} those inferred from user ratings.
Expressed preferences measure what users \textit{think} are more helpful, but to see if this agrees with what \textit{truly} helps users, we introduce \textbf{observed preferences:} those inferred from observable goals (e.g. learning) as users interact with outputs.
We collect observed preferences via the mean time users need to learn a term while studying with its mnemonic, a proxy for mnemonic short-term learning efficacy.


Over three months, 45 students gave 2684 preferences on mnemonic pairs. 
To decide how to~align \model, we study the relation of preference types.
Expressed and observed preferences disagree (\cref{section:analysis}), so what students \textit{think} helps them learn differs from what \textit{truly} helps them learn.
These preference types represent equally valuable goals (\cref{section:rlhf}): an effective and helpful mnemonic should be non-harmful and liked by users (expressed), but also aid learning~(observed).
Thus, we design a Bayesian model \cite{gelman2006data} that learns mnemonic effectiveness via feedback from all preference types (\cref{subsection:bayesian_rlhf}).
We compare mnemonics by effectiveness to elect a winning mnemonic in the pair (Figure~\ref{fig:intro}, right) and tune \model with Direct Preference Optimization \cite[DPO]{rafailov2024direct} on this signal~(\cref{subsection:DPO}).

We assess \model through several experiments.
Fine-tuning and DPO enhance \model, so aligning LLMs to student preferences improves educational text~(\cref{subsection:ablation}). 
Further, combining all preferences via Bayesian modeling can resolve ties or missing labels in the typical method of pairwise comparisons, augmenting DPO data for mnemonic quality gains (\cref{subsection:rl_eval}). 
Multiple preferences can be gathered in one app, sometimes with no extra annotations (\cref{subsubsection:obs_pref}).
So if resources allow, we advise collecting multiple preferences to study the relation of complementary alignment objectives, which can then be combined for data augmentation to improve LLM outputs. 

Lastly, two mnemonic experts assess mnemonics (\cref{subsection:qual}) from \model, GPT-4, and a freelance creative writer from Upwork, finding: 1) \model matches the SOTA LLM GPT-4, showing the utility of student feedback; and 2) Our writer's keywords are much simpler and their explanations are more imageable compared to GPT-4 and \model, motivating mnemonic generation as a difficult task~and giving insights into feedback types (simplicity, imageability) that can also be collected to better align LLMs for downstream tasks. Our contributions~are: \\
\textbf{1)} We design \model, an LLM mnemonic generator aligned by feedback from real-world students. \\
\textbf{2)} We analyze expressed and observed preferences, finding that the LLM outputs students \textit{think} help them learn is not what \textit{actually} helps them learn.\\
\textbf{3)} We align \model with multiple preferences via Bayesian modeling, which can break preference ties for DPO output quality gains and results in a more efficient keyword mnemonic generator that matches the state-of-the-art LLM GPT-4.\\
\textbf{4)} We release the first fine-tuning and preference datasets to aid research in mnemonic generation.

\section{An Initial \model Mnemonic Model} \label{section:initial_model}

Given a vocabulary (vocab) term $v$ (e.g. \textit{\textbf{Benevolent}}), we desire a keyword mnemonic $\mnemonic$ students can use to remember the meaning of $v$. 
For optimal benefits \cite{mcdaniel1984putting}, $\mnemonic$ should link to a similarly sounding and simpler keyword $k$ (e.g. \textit{Benevolent sounds like \textbf{benefit}}), and then explain how $k$ and $v$ are linked (e.g. \textit{A boss who gives their employees \textbf{benefits} is kind---or \textbf{benevolent}}).

We now train an initial \model model to generate keyword mnemonics. 
We collect a high-quality dataset of user-written mnemonics (\cref{subsection:data_collection}, \cref{subsection:good_mnemonics}) and fine-tune \model on this mnemonic dataset~(\cref{subsection:model_training}).

\subsection{Data Collection} \label{subsection:data_collection}

A dataset with vocab terms and mnemonics does not exist, so we curate new datasets to facilitate mnemonic research. 
We use vocab words from the Graduate Records Examination (\textsc{GRE}), a graduate admissions exam that students prepare for by learning hundreds of vocab terms \cite{nayak-etal-2017-v}. 
Mnemonics have been used to help students learn GRE vocabulary \cite{fairbanks1977vocabulary, pi2021instructor}.

We base our dataset on 2380 public English GRE terms $\mathcal{V}$ from seven tutoring sites \cite{vince_vocab}.
We find a mnemonic for each term from MnemonicDictionary \cite{mnemonic_dictionary}, a site where users submit keyword mnemonics for vocab terms. 
Users can also vote on mnemonics, which we later use to find high-quality mnemonics (\cref{subsection:good_mnemonics}). 
With permission from the owners, we collect 13955 candidate MnemonicDictionary mnemonics for our dataset.

\subsection{Identifying High-Quality Mnemonics} \label{subsection:good_mnemonics}

The user-submitted mnemonics collected from \cref{subsection:data_collection} are noisy, but a subset of high-quality data would better train \model for mnemonic generation \cite{xia2024less}. 
MnemonicDictionary users upvote or downvote mnemonics, so upvote ratio could find high-quality data, but this metric does not consider all upvotes given \cite{doi:10.1177/0956797617711291}.
Thus, following \citet{avg_rating}, we build a Bayesian model to learn the probability $q_{i}$ of mnemonic $\mnemonic_i$ being high-quality, based on the upvote $\nu_{u, i}$ and downvote $\nu_{d, i}$ counts on $\mnemonic_i$. 
We assume mnemonics with higher $q_{i}$ have more upvotes, so we model $\nu_{u, i}$ as a Binomial distribution with probability $q_{i}$:
\begin{align}
    q_{i} & \sim \mathrm{Beta}(\alpha=2, \beta=8) \label{eq:quality_prior}, \\
    \nu_{u, i} & \sim \mathrm{Binomial}(\nu_{u,i} + \nu_{d,i}, q_{i}),
\end{align}
\noindent which has prior $\alpha=2, \beta=8$, as our brief manual assessment found that ${\sim}20\%$ of the mnemonics are high-quality. 
We estimate $q_i$ via No U-Turn Sampling \cite[\textsc{NUTS}]{hoffman2014no}. 
Pairs ($v_i$, $m_i$) with the 1000-highest $q_i$ values form the fine-tuning dataset $\mathcal{D}_{ft}$ for \model (details in Appendix~\ref{appendix:data}). 

\subsection{Model Fine-Tuning} \label{subsection:model_training}

The dataset $\mathcal{D}_{ft}$ has term/mnemonic pairs $(v, \mnemonic)$, so we can use $\mathcal{D}_{ft}$ to train an initial seq2seq \model model $p_{0}(\mnemonic \g v)$ to create $m$ from $v$.
Upon inspection, we find some quality issues in the mnemonics, so we use GPT-4 to clean grammar errors in $\mnemonic$ via a 0-shot prompt, and discard any $\mnemonic$ with offensive or overly culturally-specific text (see Appendix~\ref{appendix:cleaning}). We end up with 889 pairs for fine-tuning \model.

Each $(v, \mnemonic) \in \mathcal{D}_{ft}$ forms prompt~$\mathcal{P} = \text{``}\texttt{Term:}$ $v \backslash \texttt{nMnemonic:}\text{''}$ and output text $m$. 
Our~initial model $p_{0}(\mnemonic \g v)$  fine-tunes LLaMA-2 70B \cite{touvron2023llama} to minimize the cross-entropy loss $\mathcal{L}_{CE}$ of predicting tokens $m_j \in m$ given $\mathcal{P}$:
\begin{equation}
    \mathcal{L}_{CE}  = \sum_{j=1}^{|\mnemonic|} \log p(\mnemonic_j \g \mnemonic_1,..., \mnemonic_{j-1}, \mathcal{P}). \label{eq:CE}
\end{equation}
\noindent We use QLoRA \cite{dettmers2023qlora} to minimize $\mathcal{L}_{CE}$. All parameters are listed in Appendix~\ref{appendix:setup}.

\section{Collecting Mnemonic Preferences} \label{section:user_study}

Only fine-tuning does not explicitly guide \model toward mnemonics that users prefer and help them learn---our overall goal.
Thus, we use alignment \cite{ziegler2019fine}: tuning LLMs to preference labels capturing which outputs users favor. 
To align \model, we need a preference dataset $\mathcal{D}_{pref}$ with entries of a term $v$, mnemonic pair $(\mnemonic_A, \mnemonic_B)$ for $v$ created by the initial model $p_0(m \g v)$, and preference label $y \in \{\texttt{A}, \texttt{B}, \texttt{tie}\}$ noting the mnemonic in the pair users favor.
To build $\mathcal{D}_{pref}$, we create mnemonic pairs (\cref{subsection:mnemonic_pairs}), define our preference labels (\cref{subsection:preference_types}), and describe our user study details (\cref{subsection:user_study_design}).

\subsection{Generating Mnemonic Pairs} \label{subsection:mnemonic_pairs}

For $\mathcal{D}_{pref}$, we first need mnemonic pairs $(\mnemonic_A$,$\mnemonic_B)$ created by model $p_{0}(\mnemonic \g v)$ for many terms $v$.
In preference datasets, researchers sample candidate LLM responses to match the abilities of the target LLM \cite{bai2022training}, so the mnemonics should have \emph{high probability} in our model $p_{0}(\mnemonic \g v)$.
Further, to ensure the mnemonics in the pair are not too similar, which would often result in preference ties, we seek a \textit{diverse} pair of mnemonics; this gives the user distinct choices, yielding clearer preferences.

We combine these two objectives through best-of-$n$ sampling \cite{nakano2021webgpt}, which samples $n$ LLM outputs and picks the one with the best score from a reward model. 
We define a reward $\pi_{pair}(\mnemonic_A, \mnemonic_B, v)$ that returns the sum of $\mnemonic_A$ and $\mnemonic_B$ sequence probabilities from $p_{0}(\mnemonic \g v)$, minus the ROUGE-1 \cite{lin2004rouge} of $\mnemonic_A$ and $\mnemonic_B$, assessing mnemonic diversity \cite{shaib2024standardizing}.
The reward favors mnemonics with high sequence probability and low word overlap.
To create mnemonic pairs, we sample 1000 terms $\mathcal{V}_{pref} \subset \mathcal{V}$ not used in $\mathcal{D}_{train}$.
For each term $v \in \mathcal{V}_{pref}$, we sample five mnemonics $\mathcal{M} = \{m\}^5 \sim p_{0}(\mnemonic \g v)$~with 0.3 temperature. 
We take $(m_A, m_B) \in \mathcal{M} \times \mathcal{M}$ with the best $\pi_{pair}(m_A, m_B, v)$ score as the pair for $v$.

\subsection{Preference Label Collection} \label{subsection:preference_types}

\begin{figure}
    \centering
    \fbox{\includegraphics[width=\linewidth]{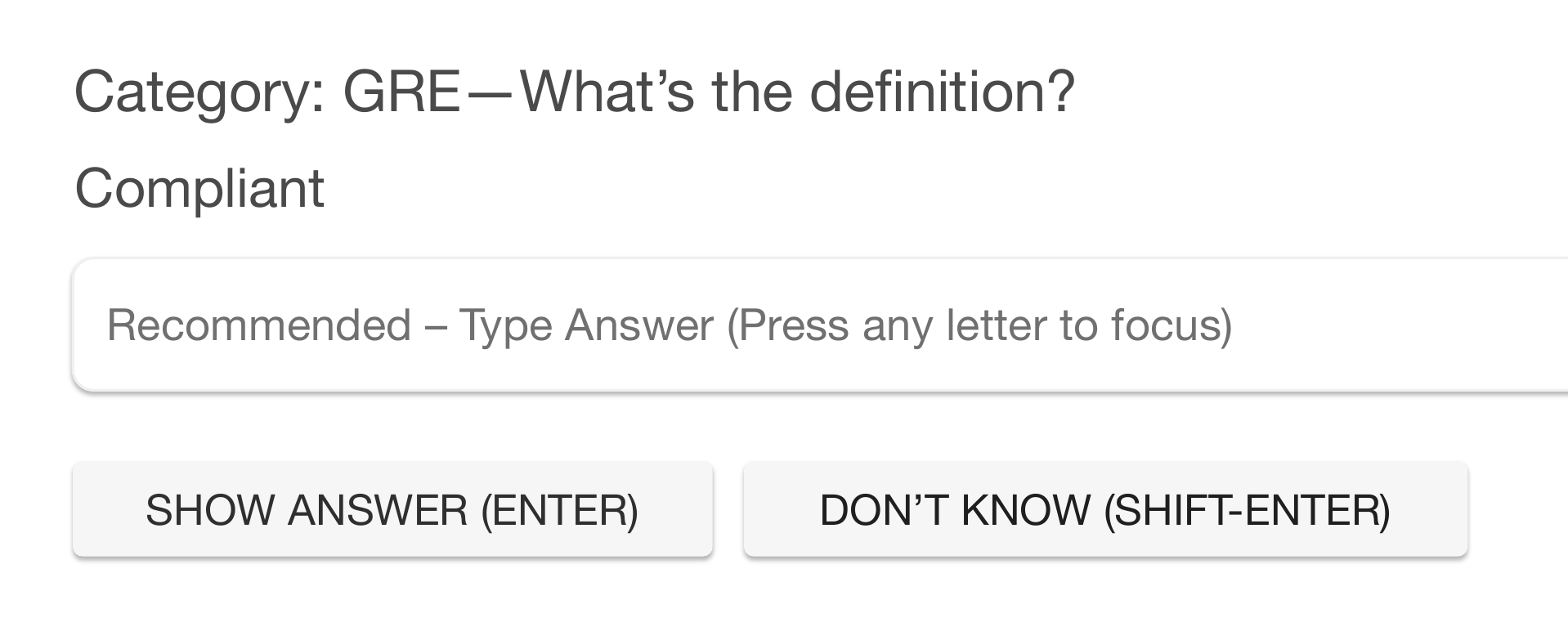}}
    \caption{Screenshot from our web-based flashcard app after a user is presented a GRE vocabulary flashcard.}
    \label{fig:vocab_ui}
\end{figure}

While preferences are often elicited through crowdworker sites, we decide to collect preferences from students who can learn vocab using our mnemonics. 
Flashcard software can aid mnemonic use \cite{picard}, so we host our user annotation schema within a flashcard app to gather student preferences $y \in \{\texttt{A}, \texttt{B}, \texttt{tie}\}$ for the mnemonic pairs in \cref{subsection:mnemonic_pairs}. 

Flashcards have two sides and while studying, users read the front of the card and answer what is on the back. In our app, users study flashcards~$f_v$ with a term $v$ as the front of the card and type its definition on the back (Figure~\ref{fig:vocab_ui}).
In one session, users study a set of flashcards with 5 to 50 terms the user has not yet studied, and continue studying until they correctly type each term's definition.
We use TF-IDF \cite{sparck1972statistical} with a cutoff of 0.15 to check if the user's typed definition matches the ground-truth definition, which the user can override if they disagree with the metric's prediction.
If a user answers $f_v$ correctly, it is removed from the pool of cards left to study.
If answered incorrectly, they~see a mnemonic from the pair (\cref{subsection:mnemonic_pairs}) for $v$ to aid learning (Figure~\ref{fig:ratings_ui}), and the card remains in the set of cards to study. 
Thus, for each card $f_v$, we can count how many turns $n \in \mathbb{Z}^+$ the user needed to answer $f_v$ correctly.
We use the \KARL \cite{shu2024karl} model and flashcard learning interface to select the set of new flashcards to show users.

Researchers often use one method to collect preferences.
But diverse methods, like pairwise comparisons and Likert ratings, can yield conflicts on which outputs are favored \cite{ethayarajh-jurafsky-2022-authenticity, bansal2023peering}, and may also give complementary signals for LLM output quality (\cref{section:rlhf}).
To see how diverse schema impact preferences in education, we collect three different preference labels grouped into two types: \textbf{expressed} and \textbf{observed}.

\subsubsection{Expressed Preferences} \label{subsubsection:verb_pref}

\begin{figure}
    \centering
    \fbox{\includegraphics[width=\linewidth]{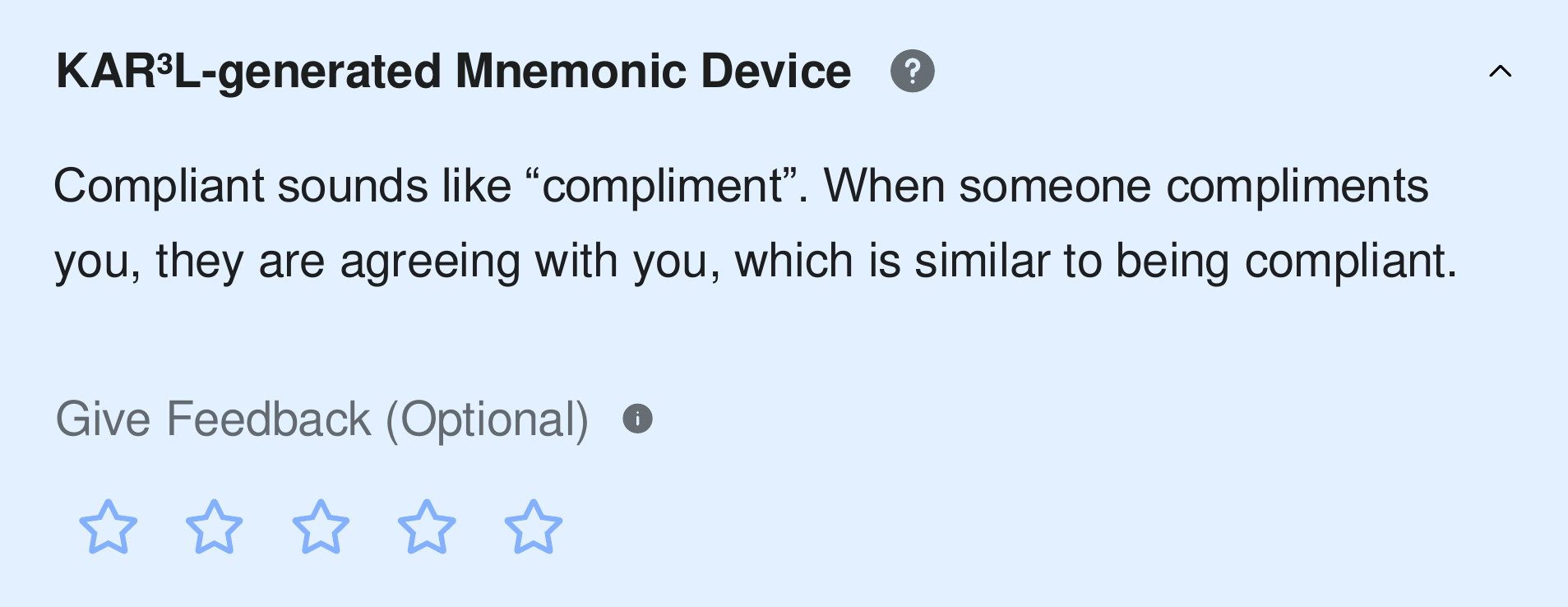}}
    \caption{Screenshot of UI to collect Likert ratings.}
    \label{fig:ratings_ui}
\end{figure}
We define \textbf{expressed preferences} as those inferred from explicit user ratings---the most common preference type \cite{casper2023open}.
We collect two expressed preferences: Likert ratings \cite{harpe2015analyze} and pairwise comparisons \cite{bozoki2013analysis}.

For Likert ratings, if a user sees a mnemonic $\mnemonic_X$ after answering card $f_v$ \textit{incorrectly}, they rate $\mnemonic_X$ on a 5-Likert scale (Figure~\ref{fig:ratings_ui}). 
We call this preference label $y_{rate}$ and set $y_{rate} = \texttt{A}$ if the average Likert rating of $\mnemonic_A$ is higher than $\mnemonic_B$ across users (same for $\texttt{B}$ and $\texttt{tie}$). 
For each $f_v$, users see only one of $\mnemonic_A$ or $\mnemonic_B$, so their rating cannot be biased by having already seen the other in the pair.
\begin{figure}
    \centering
    \fbox{\includegraphics[width=\linewidth]{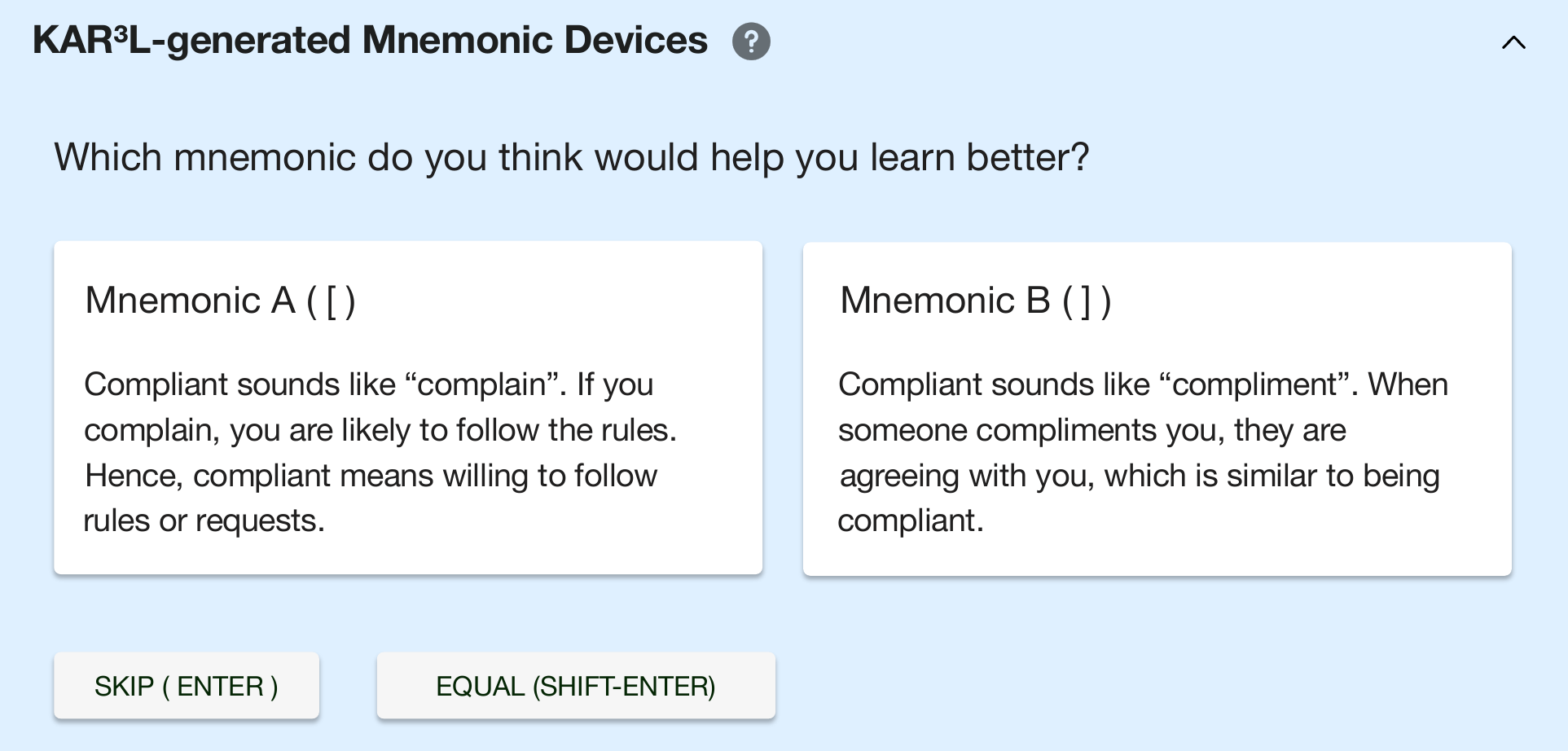}}
    \caption{Screenshot of UI for pairwise comparisons.}
    \label{fig:pairwise_ui}
\end{figure}

For pairwise preferences, if $f_v$ is answered \textit{correctly}, the user picks the mnemonic in $(\mnemonic_A, \mnemonic_B)$ they think would help them learn best (Figure~\ref{fig:pairwise_ui}). 
Users can pick one mnemonic or mark them equal.
We call this preference $y_{pair}$ and set it to the most voted option. 
Order is shuffled for position biases.

\subsubsection{Observed Preferences} \label{subsubsection:obs_pref}

Expressed preferences measure the outputs users \textit{think} are more helpful, but they do not capture what is \textit{truly} more helpful for user goals (e.g. learning).
Such preferences are undefined, so we propose \textbf{observed preferences}---those inferred from observed outcomes of user interactions with model outputs. 

As keyword mnemonics can improve short-term recall \cite{wang1992keyword}, we base observed preferences on the mean turns $t_X$ users studying with $\mnemonic_X$ need to correctly answer its flashcard~$f_v$, as calculated in \cref{subsection:preference_types}. This measure forms a proxy for short-term learning. 
For a given pair $(m_A, m_B)$, if $t_A < t_B$, we call this preference label $y_{learn}$ and set $y_{learn} = \texttt{A}$, as $\mnemonic_A$ helps users learn the definition of $v$ quicker than $\mnemonic_B$ (same for $\texttt{B}$ and $\texttt{tie}$).
$y_{learn}$ is collected automatically as users study. 


\subsection{User Study Details} \label{subsection:user_study_design}

We deploy all mnemonics from $\mathcal{D}_{pref}$ and have 47 English-speaking students from exam preparation forums, Google ads, and university courses study in our app.
To filter noisy annotators, we add random quality checks in pairwise comparisons, where one mnemonic is clearly low-quality.
If any user picks the low-quality mnemonic in the pairwise comparison, their preferences are omitted for analysis and model training.
All~users, including the two users who failed our quality checks, are awarded \$50.
In three months, 45 students gave us 2684 preferences for 752 mnemonic pairs and 472 unique pairs (Table~\ref{table:summary}, details in Appendix~\ref{appendix:summary2}). 
We omit pairs with two~or fewer labels so the mean labels per pair is 3.57, following the method of using three or more preferences to curb noise \cite{bai2022training, ji2024beavertails}.
Users are referred to just by numerical ID.



\section{Preference Analysis} \label{section:analysis}

We study the relation of our preference types (\cref{subsection:preference_types}) and uncover that students cannot fully predict what aids learning (\cref{subsection:corr}, \cref{subsection:predict_learning}).
Thus, we conclude that a mnemonic's overall helpfulness cannot be captured by just one preference type, inspiring the design of our final model that combines all preferences (\cref{section:rlhf}).
\begin{table}[t]
\small
\centering
\begin{tabular}{@{}ccc@{}}
\toprule
\textbf{Preference Pairs} & \textbf{Raw Agreement} & \textbf{Sample Size} \\ \midrule
$(y_{pair}, y_{rate})$ & 0.675 & 80 \\
$(y_{rate}, y_{learn})$ & 0.507 & 73 \\
$(y_{pair}, y_{learn})$ & 0.407 & 59 \\ \bottomrule
\end{tabular}
\caption{\label{table:correlation} Raw agreement of preference types. Expressed preferences $y_{pair}$ and $y_{rate}$ have some disagreement, but agreement between expressed and observed preferences ($y_{rate}$ vs $y_{learn}$ and $y_{pair}$ vs $y_{learn}$) is even~lower.}
\end{table}
\subsection{Are Preference Types Equivalent?} \label{subsection:corr}

To see if our preference labels capture equivalent information, we compute the agreement of preferences (e.g. $y_{pair}$ vs $y_{rate}$) for the same mnemonics.
We exclude labels denoting a tie, focusing instead on labels that show a clear preference towards one mnemonic. 
Table~\ref{table:correlation} shows that the expressed preferences $y_{pair}$ and $y_{rate}$ have moderate agreement ($0.675$), aligning with \citet{bansal2023peering}, who also uncover disagreement in pairwise and rating preferences. But notably, the agreement between expressed and observed preferences is much lower ($0.507$ and $0.407$), so asking students which outputs they think are more helpful does not always capture what is truly more helpful for learning.





\begin{figure}[t]
    \centering
    \includegraphics[width=\linewidth]{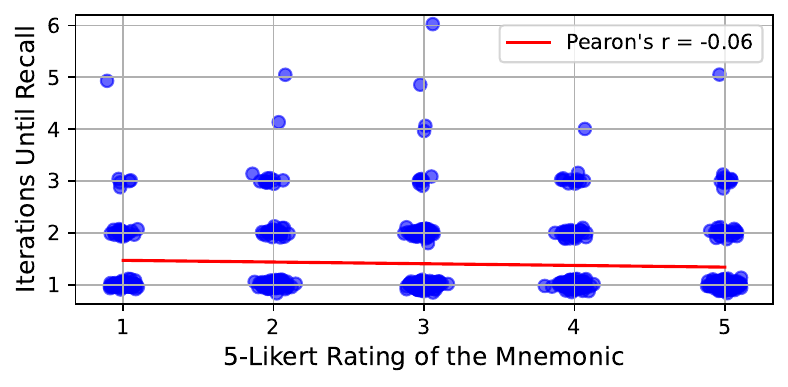}
    \caption{Correlation between user mnemonic ratings and turns needed for the same user to recall the term when studying with said mnemonic (jittered). Users~cannot predict which mnemonics will best help them learn.}
    \label{fig:pred}
\end{figure}
\subsection{Can Users Predict their Learning?} \label{subsection:predict_learning}

To study preference type disagreement at the student level, we see if a user's rating of a mnemonic predicts the total turns needed for the user to learn the vocab term linked to that mnemonic. 
If users can predict this well, we would see a strong negative correlation, with lower ratings indicating more turns needed, but we find little ($r = -0.06$) correlation (Figure~\ref{fig:pred}).
Prior work shows that students struggle to identify the \textit{study strategies} that best aid learning \cite{mccabe2011metacognitive, yan2016difficulty}, and we uncover that students struggle to do the same for \textit{study items}, further showcasing that expressed preferences cannot fully capture learning outcomes.

\section{Training a \model-er Mnemonic Model} \label{section:rlhf}

Our goal of collecting student preference labels for mnemonic pairs is to identify more \textit{helpful} or \textit{effective} mnemonics, using this signal to guide \model's outputs.
But as preference types disagree (\cref{section:analysis}), \textit{how should we identify more effective mnemonics?}

Pairwise comparisons are typically used for this purpose, but they do not always match our goal~of aiding learning (\cref{subsection:corr}).
Further, when pairwise~preferences are missing or have a \texttt{tie}, we could draw from other preferences to break~these ties and identify the better mnemonic, instead of discarding the pair for training (\cref{subsection:rl_eval}).
Conversely,~using observed preferences to select more effective mnemonics is promising as it matches our learning goal, but also using expressed preferences could help us avoid bizarre or offensive \cite{kroll1988bizarre} mnemonics that may still aid learning (see Appendix~\ref{appendix:offensive_learning}).

Given the multi-faceted nature of mnemonic effectiveness, we develop a Bayesian model to learn effectiveness from all preferences and tune \model via Direct Preference Optimization \cite{rafailov2024direct} on this learned effectiveness signal (\cref{subsection:DPO}).

\subsection{Learning Mnemonic Effectiveness} \label{subsection:bayesian_rlhf}

For mnemonic pairs $(\mnemonic_A, \mnemonic_B) \in \mathcal{D}_{pref}$, we seek to find the more effective mnemonic. We intuit that mnemonic effectiveness is a latent value that can be modeled via feedback from all three of our preferences. Bayesian models are well-suited for this task as they capture annotator noise in feedback more effectively than aggregation \cite{wang2023aligning}.

Thus, we design a Hierarchical Bayesian model \cite{gelman2006data} to estimate mnemonic effectiveness.
We seek to learn $\probP(\theta_{A} > \theta_{B})$, the probability mnemonic $\mnemonic_A$ is more effective than mnemonic $\mnemonic_B$.
To do so, we model mnemonics $\mnemonic_{A,i}$ and $\mnemonic_{B,i}$ by latent effectiveness parameters $\theta_{A,i}$ and $\theta_{B,i}$, which are assigned uniform priors:
\begin{align}
    \theta_{A,i}, \theta_{B,i} & \sim \mathrm{Beta}(1, 1). \label{eq:eff_prior}
\end{align}
\noindent We assume $\theta_{A,i}$ and $\theta_{B,i}$ influence observed feedback in our three preferences: pairwise $y_{pair}$; rating $y_{rate}$; and learning $y_{learn}$, which we outline below.

For \textbf{pairwise preferences}, let $\mathcal{C}_{i} = \{c_1, ..., c_n\}$ be the pairwise ratings of $(\mnemonic_{A, i}, \mnemonic_{B,i})$, where $c_{i} \in \{\texttt{A}, \texttt{B}, \texttt{tie}\}$.
If $\theta_{A,i} > \theta_{B,i}$, we assume $\mathcal{C}_{i}$ has more $\texttt{A}$ preferences. To model $\mathcal{C}_{i}$ from $\theta_{A,i}$ and $\theta_{B,i}$, we compute the sigmoid ($\sigma$) of a linear transform of $\theta_{A, i}$ as the probability $p^{pair}_{A, i} = \probP (\texttt{A} \in \mathcal{C}_i)$:
\begin{align}
    \alpha_{pair}, \beta_{pair} & \sim \mathrm{Normal}(0, 1), \\
    p^{pair}_{A, i} & = \sigma(\alpha_{pair} \cdot \theta_{A, i} + \beta_{pair}), \label{eq:transform}
\end{align}
\noindent and same for $p^{pair}_{B, i}$. We then model $\mathcal{C}_{i}$ as a Bradley-Terry model with ties \cite{davidson1970extending}, where $\probP(\texttt{tie} \in \mathcal{C}_i)$ depends on a uniform latent value~$\tau$: 
\begin{align}
    \tau & \sim \mathrm{Beta}(1, 1), \\
    p^{pair}_i & = \frac{[p^{pair}_{A, i}; p^{pair}_{B, i}; \tau]}{p^{pair}_{A, i} + p^{pair}_{B, i} + \tau}, \\
    \mathcal{C}_i & \sim \mathrm{Multinomial}(n, p^{pair}_i).
\end{align}
For \textbf{rating preferences}, let $\mathcal{R}_{A,i} = \{r_1, ..., r_5\}$ be cumulative counts of Likert ratings for $\mnemonic_{A, i}$, where $\mnemonic_{A,i}$ has $r_j$ votes less than or equal to rating $j$. 
We assume $\mnemonic_{A, i}$ with higher effectiveness $\theta_{A, i}$ has higher ratings.
We model $\mathcal{R}_{A,i}$ as a multinomial distribution, parameterized by a linear transformation of $\theta_{A, i}$ to a 5-length probability distribution: 
\begin{align}
    \alpha_{rate}, \beta_{rate} & \sim \mathrm{Normal}(0, 1)^5, \\
    p^{rate}_{A, i} & = \sigma(\alpha_{rate} \cdot \theta_{A, i} + \beta_{rate}), \\
    \mathcal{R}_{A,i} & \sim \mathrm{Multinomial}(\Sigma \mathcal{R}_{A, i}, p^{rate}_{A, i}),
\end{align}
and do the same for $\mathcal{R}_{B,i}$. 

For \textbf{learning preferences}, $\mathcal{T}_{A, i}=\{t_1,...,t_m\}$ is the distribution of turns users need to recall the term with $\mnemonic_{A,i}$, where $t_j \in \mathbb{Z}^+$. We assume $m_{A,i}$ with higher effectiveness $\theta_{A, i}$ yields fewer $t_j$ needed. Every turn count $t_j$ is the tries until a success, so we model each $t_j$ as a Geometric distribution parameterized by a linear transformation of $\theta_{A, i}$:
\begin{align}
    \alpha_{learn}, \beta_{learn} & \sim \mathrm{Normal}(0, 1), \\
    p^{learn}_{A, i} & = \sigma(\alpha_{learn} \cdot \theta_{A, i} + \beta_{learn}), \\
    t_j & \sim \mathrm{Geometric}(p^{learn}_{A, i}),
\end{align}
and same for $\mathcal{T}_{B, i}$. 
We learn all variables via NUTS \cite{hoffman2014no} for 1000 epochs.
Parameters converge across five chains (Appendix~\ref{appendix:bayes}), meaning our model consistently estimates effectiveness. 

\subsection{Aligning \model with Student Preferences} \label{subsection:DPO}

We now use the learned effectiveness of mnemonics $(\mnemonic_{A}, \mnemonic_{B})$ in the preference dataset $\mathcal{D}_{pref}$ to align \model.
Among many alignment methods, we adopt Direct Preference Optimization \cite[DPO]{rafailov2024direct}, which tunes LLMs to preferences without requiring explicit reward modeling or reinforcement learning steps.
Alternatives like Proximal Policy Optimization \cite{Schulman2017ProximalPO} need extensive parameter tuning and are thus harder to reproduce \cite{shengyi2022the37implementation}.


DPO requires dataset entries with a prompt $x$ and winning/losing outputs $y_w$/$y_l$, where $y_w$/$y_l$ are ``good''/``bad'' outputs for $x$.
We set $x$ to the term $v$ in $\mathcal{D}_{pref}$ with its mnemonics $(\mnemonic_A, \mnemonic_B)$ as outputs. 
The mnemonic with higher effectiveness ($\theta_A$ vs $\theta_B$ from \cref{subsection:bayesian_rlhf}) is $y_w$, and the other is $y_l$.
With this data, we update our initial model $p_{0}(m \g v)$ ($\pi_0$ below) to align a better \model model $p_{dpo}(m \g v)$ ($\pi$ below) with DPO, which minimizes the loss~$\mathcal{L}_{dpo}$:
\begin{equation} \small
\mathcal{L}_{dpo} = \mathop{-\mathbb{E}}_{\substack{x, y_w, y_l \\ \sim \mathcal{D}_{pref}}} \left[ \ln \sigma \left( \beta \ln \frac{\pi(y_w | x)}{\pi_{0}(y_w | x)} - \beta \ln \frac{\pi(y_l | x)}{\pi_{0}(y_l | x)} \right) \right].
\end{equation}

\noindent \model minimizes $\mathcal{L}_{dpo}$ using QLoRA~\cite{dettmers2023qlora}. Appendix~\ref{appendix:setup} lists all parameters.

\section{How Smart Are \model's Mnemonics?} \label{section:evaluation}
We now assess \model's mnemonics for 500 terms $\mathcal{V}_{test} \subset \mathcal{V}$ not used in $\mathcal{D}_{train}$ or $\mathcal{D}_{pref}$. 
\model is aligned using a combination of three preference metrics: pairwise comparisons, Likert ratings, and learning. 
Due to space and data limits, we mainly evaluate via the most popular of the three metrics: pairwise comparisons \cite{casper2023open}.
Thus, our evaluation reveals how using multiple preference labels (\textsc{MPLs}) affects \textit{pairwise} output quality.
We acknowledge that an evaluation across all preference metrics would be insightful (\cref{section:limitations}) and hope future works extend this direction with our datasets.


Given the costs of human pairwise evaluations, we adopt a common practice having GPT-4 judge which of two model-created mnemonics is higher quality \cite{chiang2023vicuna, liu-etal-2023-g}.
GPT-4 has 80\% agreement with users (Appendix~\ref{appendix:clf}) on 200 held-out mnemonic pairs, near the 81\% human agreement in MT-bench \cite{zheng2024judging}, so GPT-4 agrees with user pairwise mnemonic ratings.
To curb position bias \cite{wang2023large}, we compare mnemonics in both orders, only marking that one model wins if GPT-4 picks the model's mnemonic in both orders, otherwise marking a tie.

We first use GPT-4 to compare mnemonic quality of \model ablations (\cref{subsection:ablation}, \cref{subsection:rl_eval}).
We then have mnemonic experts evaluate \model's mnemonics to inform future work~(\cref{subsection:qual}).
We also present examples of \model's mnemonics in Appendix~\ref{appendix:examples}.


\subsection{Ablation Study} \label{subsection:ablation}

We ablate \model (Figure~\ref{fig:intro}) to verify that both fine-tuning and DPO improve mnemonic quality.
Our fine-tuned model~$p_{0}(m \g v)$ generates higher-quality mnemonics versus the few-shot LLaMA model $p_{fs}(m \g v)$ prompted using the ten $\mathcal{D}_{ft}$ examples with the highest latent quality (\cref{subsection:good_mnemonics}), and same for $p_{dpo}(m \g v)$ versus $p_{0}(m \g v)$ (Table~\ref{table:ablation}).
Both of the steps improve pairwise mnemonic quality, confirming DPO can align LLMs with student preferences to enhance LLM outputs in education.

\subsection{DPO with Multiple Preference Labels} \label{subsection:rl_eval}

We investigate the effectiveness of training with DPO using MPLs for pairwise mnemonic quality through two research questions, outlined below: \\

\noindent \textbf{Q1---Do MPLs harm pairwise metrics?}
One concern of optimizing on MPLs with DPO is that the model will produce lower-quality mnemonics compared to a model using pairwise labels, as the latter optimizes just on the evaluation metric.
To test this, we first select a subset of preference data $\mathcal{D}_{pair} \subset \mathcal{D}_{pref}$\footnote{To have enough label disagreement, $\mathcal{D}_{pref}$ in this analysis also adds mnemonics with two labels (see Appendix~\ref{appendix:dataset_splits}).} with the pairwise preference $y_{pair}$.
We then train two DPO models on $\mathcal{D}_{pair}$ when $y_{pair} \neq \texttt{tie}$: $p_{bayes}(m \g v)$, training on the Bayesian label $y_{bayes}$ from \cref{subsection:bayesian_rlhf}, and $p_{pair}(m \g v)$, training on $y_{pair}$. 
Despite having 20\% disagreement in $y_{pair}$ and $y_{bayes}$ on the winning mnemonic, the two models are judged to generate mnemonics with equal quality on $\mathcal{V}_{test}$ (Table~\ref{table:quant}, top). 
Thus, DPO training with MPLs does not always degrade LLM output quality on singular preference metrics. \\


\begin{table}[]
\small
\centering
\setlength{\tabcolsep}{10pt}
\begin{tabular}{@{}cccc@{}}
\toprule
\textbf{Model A/B Pair} & \textbf{A Wins} & \textbf{Tie} & \textbf{B Wins} \\ \midrule
$p_0(m | v)$, \; $p_{fs}(m | v)$ & \textbf{0.76*} & 0.13 & 0.11 \\
$p_{dpo}(m | v)$, \; $p_{0}(m | v)$ & \textbf{0.29*} & 0.53 & 0.18 \\ \bottomrule
\end{tabular}
\vspace{-1.5ex}
\caption{GPT judgement of our ablations. Significantly better models (Binomial, $p<0.005$) are \textbf{bold} with~*. Our fine-tuning and DPO steps both improve \model.}
\label{table:ablation}
\end{table}
\begin{table}[]
\small
\centering
\setlength{\tabcolsep}{7pt}
\begin{tabular}{@{}cccc@{}}
\toprule
\textbf{Model A/B Pair} & \textbf{A Wins} & \textbf{Tie} & \textbf{B Wins} \\ \midrule
$p_{bayes}(m | v)$, \; $p_{pair}(m | v)$ & 0.19 & 0.60 & 0.21 \\
$p_{dpo}(m | v)$, \; $p_{pair}(m | v)$ & \textbf{0.28*} & 0.54 & 0.18 \\ \bottomrule
\end{tabular}
\vspace{-1.5ex}
\caption{GPT judgement of DPO models. Significantly better models (Binomial, $p<0.005$) are \textbf{bold} with~*. Multiple preferences can break ties in singular preferences for mnemonic quality gains (bottom row).}
\label{table:quant}
\vspace{-0.9ex}
\end{table}


\noindent \textbf{Q2---Can MPLs augment data?} If we train~DPO with just $y_{pair}$, we must discard preference data when $y_{pair} = \texttt{tie}$ or no $y_{pair}$ exists.
While we could collect more $y_{pair}$ labels with another user study, we investigate whether $y_{bayes}$ can directly resolve the missing or tied $y_{pair}$ labels using the other preferences to elect winning mnemonics, augmenting our training data without collecting any more pairwise comparison data.
We compare $p_{pair}(m \g v)$, which trains on 348 $y_{pair}$ labels, to our full model $p_{dpo}(m \g v)$, which trains on 117 extra pairs when $y_{bayes}$ breaks a tie in $y_{pair}$ and twelve extra pairs without $y_{pair}$ labels.
$p_{dpo}(m \g v)$ has significantly better ($p<0.005$) mnemonics than $p_{pair}(m \g v)$ (Table~\ref{table:quant}, bottom), meaning that MPLs can effectively augment DPO training data over pairwise preferences for output quality gains.

\noindent \textbf{Takeaway:} Since optimizing on $y_{bayes}$ matches $y_{pair}$ in non-ties and improves output quality by resolving ties and missing $y_{pair}$ labels, we advise collecting MPLs if resources allow.
This is feasible as MPLs can be collected in a single app, which is often cheaper than another user study to break ties, especially as some labels (e.g. $y_{learn}$) can~be gathered without explicit annotations.
Such efforts can help researchers study complementary alignment objectives (\cref{section:analysis}) and even boost LLM output quality.

\subsection{Qualitative Evaluation} \label{subsection:qual}

For a detailed evaluation, we have two mnemonic researchers assess our mnemonics, split into keyword and explanation quality. 
For \textit{keyword quality}, we ask two yes/no questions:
1) Does the keyword sound like the term? (\textbf{P}honetic \textbf{S}imilarity);
2) Is the keyword simpler\footnote{A keyword is simpler if a user not knowing the term~could likely know the keyword (\emph{torpor} is not simpler than~\emph{torpid}).} than the term? (\textbf{Simplicity}). For \textit{explanation quality}, we rate mnemonic explanations out of five on:
1) \textbf{Clarity:} Ease of understanding;
2) \textbf{Str}ength: The obviousness of the explanation's association of the keyword and the term \cite{hall1981mnemotechnics};
3) \textbf{Image}ability: The ability to evoke mental imagery \cite{campos2011using}.
We~argue clarity and strength are most important for explanation quality, ensuring students can understand and create strong memory links from terms to keywords.
We use imageability as it \textit{can} affect memory \cite{groninger1971mnemonic}, but an imageable explanation is still unmemorable if it is unclear or low strength.\footnote{For example, \emph{Ben Franklin} for \emph{benevolent} in Figure~\ref{fig:intro}.}

We assess mnemonic keyword and explanation quality for 50 $\mathcal{V}_{test}$ terms created by a professional writer from Upwork, 10-shot GPT-4, and \model, $p_{dpo}(m \g v)$.
We also compare \model's keyword quality to Transphoner \cite{savva2014transphoner}, a SOTA mnemonic keyword extractor. 
Transphoner is hard to reproduce, so we use 50 terms and Transphoner outputs released by the authors for this comparison. \\

\begin{figure}[t]
    \centering
    \includegraphics[width=\linewidth]{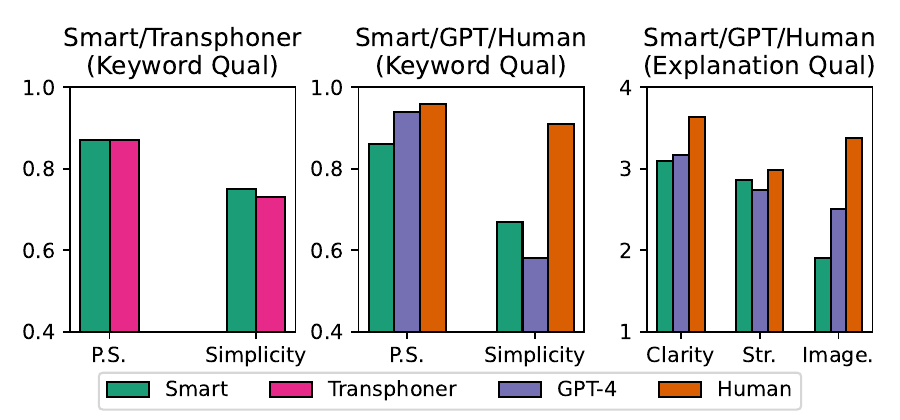}
    \vspace{-4ex}
    \caption{Expert qualitative evaluation of mnemonics. \model matches GPT-4 at much lower deployment costs, but our human writer largely surpasses all models, especially in keyword simplicity and imageability.}
    \label{fig:human}
\end{figure}

\noindent \textbf{Keyword Quality:} \model has slightly better keywords than Transphoner (higher simplicity, equal PS), meaning LLMs are strong alternatives for~keyword extractors (Figure~\ref{fig:human}, left).
\model also produces much simpler keywords than GPT-4, but with lower PS (Figure~\ref{fig:human}, middle).
\citet{suvarna2024phonologybench} reveals a large gap (37\% accuracy) when prompting LLaMA-2 13B and GPT-4 for rhyme generation which also assesses PS; our gap is just 8\%, so fine-tuning and human feedback can help LLMs address phonetic weaknesses.
Lastly, our human writer surpasses both models in keyword quality, with large simplicity gaps.
Thus, systems that simplify text with LLMs (summarizers, topic models) may benefit from explicit feedback on word simplicity. \\

\noindent \textbf{Explanation Quality:} \model matches GPT-4's explanations (lower clarity, but higher strength), meaning student feedback lets smaller LLMs like \model compete with large SOTA LLMs (Figure~\ref{fig:human}, right).
Our writer again has the best explanations, especially in imageability over \model, as \model is not explicitly guided toward imageable outputs.
Imageability does not entail memorability, but it may be useful for memorability tasks (story-telling, advertising) to use imageability feedback, as even GPT-4 shows a clear weakness in imageability. \\

\noindent \textbf{Takeaway:} Training LLMs with student feedback results in mnemonics with keyword quality matching SOTA extractive methods while also generating explanations. It also allows \model, a smaller LLM,~to compete with GPT-4.
This is a significant feat, as \model can provide students with mnemonics as effective as GPT-4 at much lower deployment costs---an attractive feature for educators looking to implement LLM educational tools. 
Further, since GPT-4 is stronger than LLaMA, using student feedback with GPT-4~could yield even better mnemonics.
Lastly, our expert writes better mnemonics~than LLMs, specifically in keyword simplicity and imageability. This motivates mnemonic generation as a challenging task and gives insights into feedback that could be used to guide LLMs in similar tasks.

\section{Related Work}

Below, we review relevant literature on mnemonic generation (\cref{subsection:mnemonic_rw}) and human preferences (\cref{subsection:preferences_rw}).

\subsection{Mnemonic Generation} \label{subsection:mnemonic_rw}

Mnemonics help users learn information, such as passwords \cite{yang2016empirical, song2019alphapwd}, vocabulary~\cite{dundes1961mnemonic, levin1992mnemonic}, and medical facts \cite{ajayi2019feasibility, leeds2020teaching}.
There are many mnemonic types, including song \cite{hayes2009use, werner2018music}, acronym \cite{mccabe2013psychology, li2021psg}, and keywords \cite{campos2004importance, campos2011using}. 
We study keyword mnemonics, which link complex terms to simpler keywords.

The effort of manually writing mnemonics has led to mnemonic generation research. 
Early works use phonetic similarity and multi-score ranking to find keywords \cite{savva2014transphoner, anonthanasap2015imnem, 7409024}, but these methods do not explain how the keyword is linked to the fact. 
Recent works prompt LLMs to generate mnemonic explanations \cite{lee2023smartphone, wong2024large}, but we are the first to collect fine-tuning and preference data to generate mnemonics guided by real-world student~feedback. 

Prior education work has found that students cannot predict which study strategies, such as blocked versus interleaved practicing, will best help them learn \cite{mccabe2011metacognitive, yan2016difficulty}.
We discover that students struggle to do the same for individual study items (\cref{subsection:predict_learning}), such as mnemonics.

\subsection{Human Preferences} \label{subsection:preferences_rw}

Recent work aligns LLMs with preference data capturing what humans prefer \cite{abc}.
Alignment methods include reinforcement learning with reward models \cite{christiano2017deep, ziegler2019fine}, selecting high-quality data \cite{sanh2022multitask, zhou2024lima}, and augmenting LM loss with rewards \cite{yuan2023rrhf, rafailov2024direct}.
Preferences have been used for sentiment generation \cite{maas-etal-2011-learning}, summaries \cite{volske-etal-2017-tl}, and dialogue safety \cite{bai2022training}, but we are the first to study them in mnemonics.


Our work also follows recent efforts to measure issues in preferences, such as preference agreement \cite{ethayarajh-jurafsky-2022-authenticity, bansal2023peering} and annotator biases \cite{peng2022investigations, wan2023poisoning}. 
In contrast, we distinguish between expressed and observed preferences and show expressed preferences do not capture what truly helps users.
We are also similar to works that collect preferences across varied demographics \cite{kirk2024prism} and use reinforcement learning regularization to align models to diverse preferences \cite{xue2023reinforcement}, but we are the first to collect diverse preferences in an education setting and combine them via Bayesian modeling \cite{yang2024bayesian}.

In social science, several works find that a human's stated preferences, elicited from survey responses, do not always agree with revealed preferences, the human's actual behavior \cite{urama2006stated, hoderlein2014revealed, de2021stated}.
We show that the same applies in education, as pairwise comparisons and Likert ratings for perceived learning efficacy (expressed preferences) have low agreement (\cref{subsection:corr}) with what truly helps users learn (observed preferences).
\citet{mozannar2024realhumaneval} find a similar trend in coding, where programmer preferences do not correlate with their actual productivity.
As a result, we hope our work will lead researchers to reevaluate how we should measure helpfulness in preference data collection to design models that truly help users downstream.

\section{Conclusion}

We design \model, the first keyword mnemonic generator guided by student feedback.
\model~is trained on new fine-tuning and preference datasets, both of which are released.
While curating data, we reveal low agreement in \textbf{expressed} preferences and our introduced \textbf{observed} preferences, showing that students cannot predict their learning.
Combining expressed and observed preferences via DPO and Bayesian modeling yields a smaller, more efficient mnemonic model matching GPT-4.
However, our human writer surpasses both models, especially in keyword simplicity and explanation quality, motivating mnemonic generation as a challenging~task.
To further challenge LLMs, researchers could explore personalizing mnemonics for students, adapting \model to different domains, languages, and modalities, or optimizing on other learning signals (e.g. \textit{long-term} learning) via student feedback.

While expressed preferences are currently used for alignment, they do not measure what truly helps users.
Thus, there is a need for more work in capturing observed preferences and user goals.
Ensuring LLMs pursue such goals safely requires alignment methods to steer LLMs toward \textit{both} expressed and observed preferences, and we design a method to combine them; our method also resolves ties and missing labels to augment datasets. 
We hope our study of the disagreements and benefits of diverse mnemonic preferences will motivate future work in safely aligning LLM outputs to true user needs.


\section{Limitations} \label{section:limitations}

One limitation is that our fine-tuning and preference datasets are relatively small.
Despite this, our datasets both improve the quality of mnemonics from \model (\cref{subsection:ablation}), following the recent paradigm of LIMA \cite{zhou2024lima} which suggests that small, high-quality datasets can be used to align and improve LLMs.
Further, regardless of size, our preference dataset results in an insightful analysis of the relation between expressed and observed preferences (\cref{section:analysis}).
If more students study vocab in our app, we will update and release both of our mnemonic datasets accordingly, resulting in larger datasets to facilitate mnemonic research. 

GPT-4 as a judge can result in biases.
We used GPT-4 due to the high cost of human annotations, which has also become a standard practice \cite{chiang2023vicuna, liu-etal-2023-g, touvron2023llama, chiang-lee-2023-large, dettmers2023qlora}.
While using GPT-4, we adopt best practices for robust evaluation: 1) only evaluating on \model to avoid self-recognition \cite{panickssery2024llm}; 2) using DSPy to limit the sensitivity of prompt perturbations \cite{khattab2023dspy}; 3) ensuring GPT-4 has high agreement with humans on held-out mnemonic comparisons \cite{zheng2024judging}; and 4) running inference on both orders of mnemonic pairs to curb position bias \cite{wang2023large}.

Lastly, our final model for \model optimizes on a combination of multiple preference labels: pairwise comparisons, Likert ratings, and learning.
While training and evaluating DPO models on all three objectives independently would be insightful, we are limited by our data.
Thus, we focus on the most popular of the three---pairwise preferences---and our evaluation reveals how optimizing with multiple preference labels can impact metrics based on just one preference label.
We encourage future work to use our preference data to explore how optimizing on multiple preferences impacts metrics based on each type of preference. 
To further motivate works in this direction, we have mnemonic experts evaluate \model's mnemonics to inform which types of feedback (e.g. keyword simplicity, explanation imageability) could be collected and trained with to improve mnemonics (\cref{subsection:qual}).

\section{Ethical Considerations}

While optimizing directly on observed preferences for downstream applications is promising, we advise researchers to take caution for deploying such models in a way that impacts users.
Even our relatively harmless objective of optimizing on learning can result in bizarre mnemonics that perpetuate harms (Appendix~\ref{appendix:offensive_learning}), and we imagine that in other domains, these consequences could be more severe.
For example, training an LLM in the news domain optimized on user clicks can result in misinformation \cite{milano2020recommender}, while optimizing directly on time spent looking at an advertisement can yield harmful, addictive content.

We took precautions to avoid these harms, such as filtering out offensive mnemonics before fine-tuning (\cref{subsection:model_training}), and our protocols were approved by an Institutional Review Board to mitigate any risks during our user study.
Thus, while we advocate for research in observed preferences for LLM alignment, we also urge researchers to consider the consequences of optimizing on these objectives before deployment in user-facing applications.

\section*{Acknowledgements}

We would like to thank members of the CLIP lab at the University of Maryland and external collaborators for their discussions of this work, including David Martinez, Naina Balepur, Sandra Sandoval, Yoo Yeon Sung, and Rachel Rudinger.
We also thank GregMat, GMAT Club, and Memli App for their feedback on our user study design and for allowing us to use their testing materials in our research.
This material is based upon work supported by the National Science Foundation under Grant No. \abr{iis}-2403436 (Boyd-Graber) and \abr{dge}-2236417 (Balepur).
Any opinions, findings, and conclusions or recommendations expressed
in this material are those of the author(s) and do not necessarily
reflect the views of the National Science Foundation.
Annotations and cloud computing resources were made possible by a gift from Adobe Research.
%


\bibliography{custom}
\bibliographystyle{acl_natbib}

\clearpage

\appendix
\section{Dataset Details} \label{appendix:data}

\subsection{Cleaning Noisy Mnemonics} \label{appendix:cleaning}

\textbf{\textcolor{red}{Warning: This subsection contains an example of an offensive mnemonic.}}

We use Bayesian modeling to obtain a high-quality subset of student-submitted mnemonics on MnemonicDictionary for fine-tuning (\cref{subsection:good_mnemonics}), but we still find grammar mistakes in the mnemonics, and some of the mnemonics can be considered offensive or too culturally specific.
To fix the grammar issues, we ask GPT-4 (web interface)  to ``Fix the spelling and grammar mistakes in these mnemonics [paste 50 mnemonic devices]''.
This converts mnemonics like ``when two gender's male \& female end up together .. they produce or give rise to a CHILD..'' to the more structured form ``When two genders, male and female, come together, they produce or give rise to a child.''

We also manually remove the 111 mnemonics we thought could be harmful to users or less understood due to being too specific to being related to a certain culture.
For example, we consider the mnemonic: \textit{Spurn sounds like 's + porn.' The mnemonic implies disdainful rejection of inappropriate content.} offensive, while the mnemonic: \textit{Glower can be associated with Gulshan Grover, imagining him glaring angrily at the hero} is too culturally specific.
In future works on mnemonic generation, it could be interesting to personalize the generated mnemonics through aspects like culture.

\subsection{$\mathcal{D}_{ft}$ and $\mathcal{D}_{pref}$ Dataset Details} \label{appendix:summary2}

In Tables~\ref{table:train} and \ref{table:pref}, we provide descriptions of the columns in $\mathcal{D}_{ft}$ and $\mathcal{D}_{pref}$, respectively.
For $\mathcal{D}_{pref}$, we filter out terms and mnemonics with less than two annotations, resulting in the summary statistics described in \cref{subsection:user_study_design}, and summarized in Table~\ref{table:summary}.
Along with the filtered subset, we will release the entire dataset of human preferences. We will also continue to release data if users study vocabulary with mnemonics in our app.
The list of 500 vocab terms $\mathcal{V}_{test}$ used for testing will also be released.

Finally, in Table~\ref{table:noise}, we quantify the noise of annotator ratings in $\mathcal{D}_{pref}$. We use average entropy and variance as each instance in our dataset can be annotated by a different number of annotators, and there is no guarantee that the same annotator will be rating each of the mnemonics. We find that the average variance and entropy of our annotations are significantly lower than random chance.

Our datasets are based on publicly available GRE vocab words and mnemonics from MnemonicDictionary, and both were used as intended by the original authors.
None of our datasets contain personal information, as users are referred to just by ID.
All collected mnemonics are in English.

\subsection{Experiment Dataset Splits} \label{appendix:dataset_splits}

When conducting the analysis of human preferences and training most DPO models, we filter all mnemonic pairs with less than two human preference annotations to form $\mathcal{D}_{pref}$.
For the pairwise preferences, this means that we filtered mnemonics where the sum of votes for $\texttt{A}$ and $\texttt{B}$ was less than or equal to two.
However, for the comparison between $p_{bayes}(m | v)$ and $p_{pair}(m | v)$ in \cref{subsection:rl_eval}, we also use mnemonic pairs with exactly two votes.
We did this because otherwise, the Bayesian label $y_{bayes}$ and pairwise label $y_{pair}$ had very high agreement in which mnemonic in the pair was winning, so we would not be able to meaningfully study the differences between optimizing on just pairwise or multiple preferences.
Essentially, our filtering strategy removed the annotator noise in the pairwise preferences, so the clear pairwise preference towards one mnemonic caused $y_{bayes}$ to nearly always be in agreement with $y_{pair}$.
However, by adding back noisy annotations, we were able to better study the differences between $y_{bayes}$ and $y_{pair}$, as these two labels ended up with 20\% disagreement.

\section{Model and Experiment Details} \label{appendix:model}

\subsection{Training Setup} \label{appendix:setup}

All of our models are trained for a maximum of 24 hours using eight NVIDIA RTX:A5000 GPUs. In practice, we find that both fine-tuning and Direct Preference Optimization (DPO) converge in around 6 hours. Parameters were manually selected. Fine-tuning and DPO were both implemented using the \texttt{trl} library\footnote{\url{https://huggingface.co/docs/trl}} using a 90/10 train/evaluation split on their respective datasets. Both fine-tuning and DPO use QLoRA with LLaMA-2 (70B) with 8-bit quantization, $\alpha=32$, a dropout of 0.05, a bias of 0, and update the default attention query and value projection layers. We use 5 training epochs for fine-tuning. We perform DPO with a training batch size of 1, a beta value of 0.1, a maximum prompt length of 16 tokens, a maximum output length of 64 tokens, and use the accuracy of the reward model as the metric for the best model. We use 5 training epochs for each DPO model (Table~\ref{table:quant})
All unspecified hyperparameters are the default values.
All evaluations are from a single run.

\subsection{Decoding Strategy} \label{appendix:decoding}

We implement greedy decoding (no sampling) when generating mnemonics in \cref{section:evaluation}. For an input vocabulary word $v$, we generate its mnemonic with the prompt: $\texttt{\#\#\# Term:} v \backslash\texttt{n\#\#\# Mnemonic:}$ $v \texttt{ sounds like}$ to ensure the mnemonic follows the two-step process of: 1) linking $v$ to a simpler keyword $k$; and 2) generating an explanation that rationalizes how $v$ and $k$ are connected.

\subsection{GPT-4 Classifier Implementation} \label{appendix:clf}

Our classifier to judge the quality of mnemonics is implemented with DSPy\footnote{\url{https://dspy-docs.vercel.app/}} \cite{khattab2023dspy} and based on \texttt{gpt-4-turbo-2024-04-09}. 
To train and evaluate this classifier, we select 250 pairs of mnemonics for the same vocabulary term from MnemonicDictionary with the highest difference in our latent quality scores (\cref{subsection:good_mnemonics}), indicating a set of mnemonics with clear human preferences.
We use 25 random training examples and 25 random validation examples to optimize this prompt with DSPy and after optimization, we run inference with the classifier on a held-out set of the 200 remaining examples.
The prompt is optimized using bootstrap few-shot with random search with a maximum of 3 bootstrapped demos, a maximum of 3 labeled demos, and 10 candidate programs.
As features, the classifier uses the two mnemonics to choose from, the vocab term, and a sample sentence containing the vocabulary word from \texttt{WordsAPI}\footnote{\url{https://www.wordsapi.com/}}.
This choice of inputs was selected by assessing validation set accuracy while adding different vocabulary features, including the definition of the word, synonyms, antonyms, and part-of-speech information.
The instruction given in the DSPy signature is: \textit{Given a vocabulary term, a sentence using the term, and two candidate mnemonics (Mnemonic A and Mnemonic B), classify whether Mnemonic A or Mnemonic B is a better mnemonic device. Output just the letter of the better mnemonic ("A" or "B")}.
Our classifier prompt will be released.

\subsection{Bayesian Model Evaluation} \label{appendix:bayes}

\begin{figure*}
    \centering
    \includegraphics[width=0.7\linewidth]{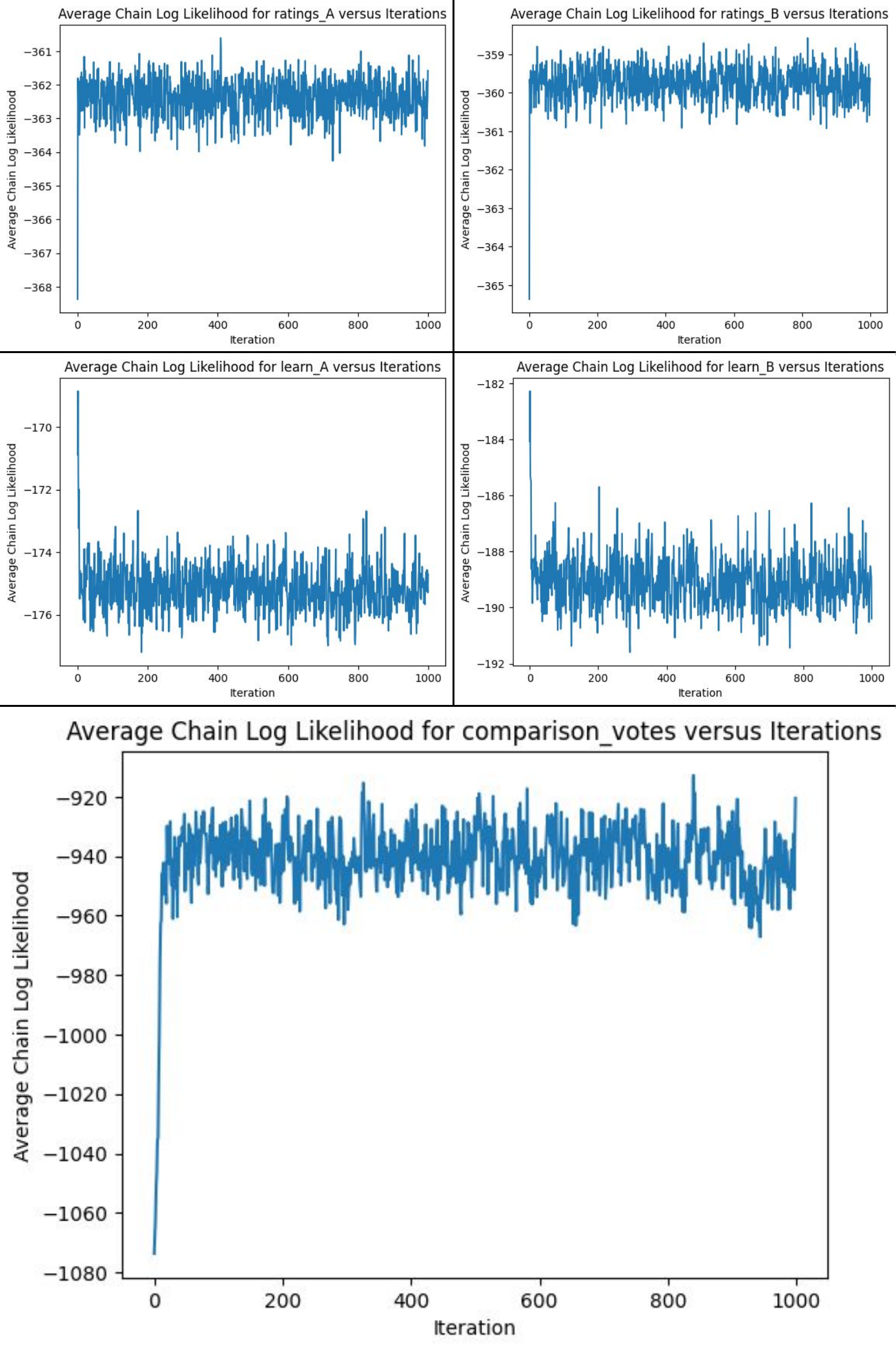}
    \caption{Log likelihood convergence of observed data in our Bayesian model.}
    \label{fig:loss}
\end{figure*}

In this section, we evaluate the quality of our Hierarchical Bayesian model.
Since our goal was to estimate the latent effectiveness of mnemonics, we first assess the convergence of our learned parameters across chains. 
All learned parameters have an $\hat{r}$ under 1.01 and an effective sample size over 1000, indicating strong convergence.
Further, our final Bayesian preference label (which mnemonic has a higher effectiveness score) across chains has a Krippendorff's $\alpha$ over 0.75, indicating strong agreement and convergence.
Finally, in Figure~\ref{fig:loss}, we display the log-likelihood values for our observed data across iterations, finding that they converge.

\begin{figure*}
    \centering
    \includegraphics[width=0.7\linewidth]{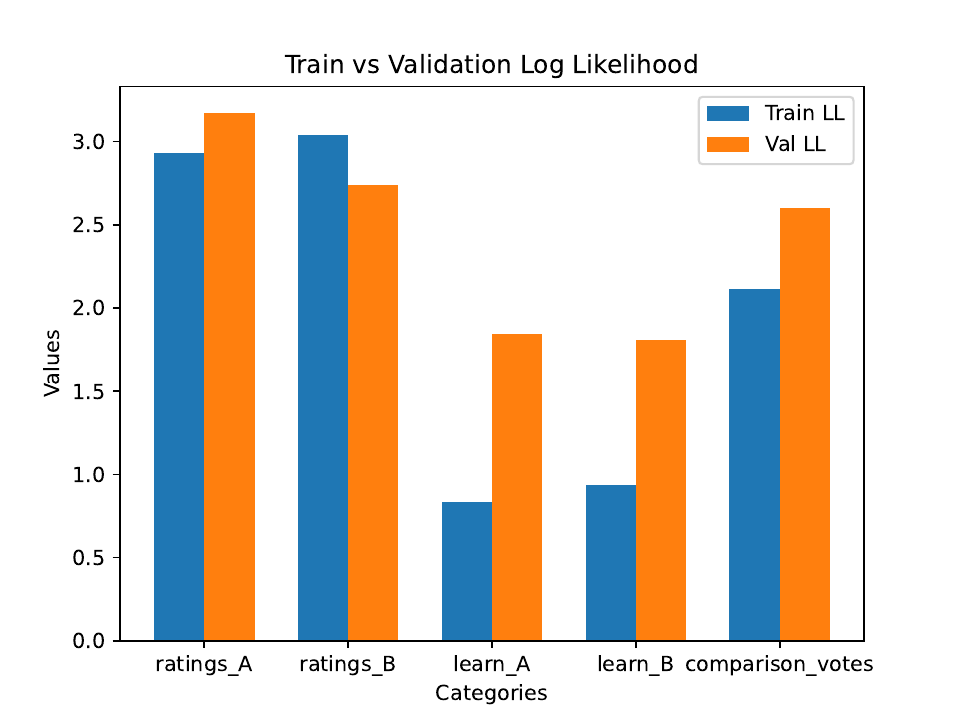}
    \caption{Comparison of Bayesian model log likelihood of observed data on training versus validation set.}
    \label{fig:generalizability}
\end{figure*}

While convergence is more important to assess the quality of a Bayesian model that learns latent values, we also assess the generalizability of our model.
We first train our model on 80\% of our data and run inference on the remaining 20\% for evaluation.
In Figure~\ref{fig:generalizability}, we compare the log-likelihood of predicting the observed data on the training and evaluation splits.
For 3/5 of our observed data types, we find a non-significant difference between the log-likelihoods (2-sample $t$-test).
The only significant difference is in the data associated with the learning preferences $y_{learn}$, further suggesting that modeling observed preferences is a challenging and interesting direction for future research.

\subsection{Obtaining Bayesian Preference Labels} \label{appendix:bayes2}

To get the final latent overall mnemonic effectiveness values for training the DPO model $p_{dpo}(m | v)$, we average the latent variables over all five chains post-burn-in. 
For the ablation study where we compare optimizing on $y_{pair}$ versus $y_{bayes}$, we similarly obtain the final latent mnemonic effectiveness values by averaging the latent variables over all five chains post-burn-in, but this time we just use a random sample of the epochs (i.e. thinning). By taking a sample of each chain instead of using the entire chains, we introduce more variability and disagreement in the labels, allowing us to more meaningfully study the difference between optimizing on pairwise preferences versus all preference labels.

\section{User Study Details} \label{appendix:user_study}

\subsection{Annotator Instructions} \label{appendix:instructions}

\begin{figure*}
    \centering
    \fbox{\includegraphics[width=\linewidth]{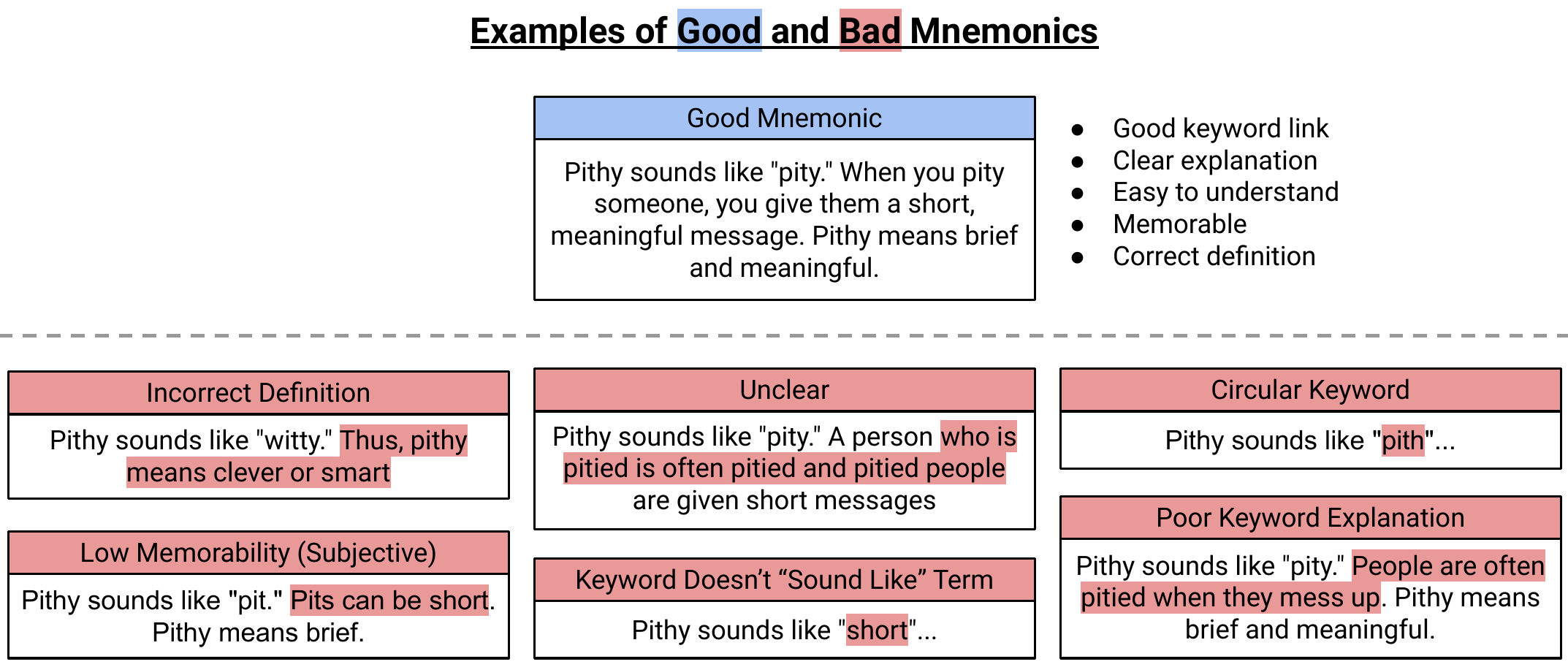}}
    \caption{Instructions given to students to help them rate the quality of mnemonic devices for expressed preferences.}
    \label{fig:instructions}
\end{figure*}

During our user studies, we ensure to provide ample instructions to annotators.
On the home page of our flashcard learning app, users can view our Institutional Review Board documents, which detail the purpose of the user study and how user data will be collected and used.
Further, we provide users instructions to help them rate the quality of mnemonic devices (Figure~\ref{fig:instructions}), which can be viewed at any time throughout the user study.
All annotators are English speakers.
Users were aware that they were participants in a research study and as participation was voluntary, compensation is fair.

\subsection{Qualitative Evaluation Details} \label{appendix:qual}

In our qualitative evaluation, we compare our full model trained with DPO on Bayesian labels versus 10-shot GPT-4 (\texttt{gpt-4-turbo-2024-04-09}), where examples were chosen according to the highest latent quality scores in $\mathcal{D}_{train}$ (\cref{subsection:good_mnemonics}). We also compare against expert human-written mnemonics to serve as an upper bound on mnemonic quality. These mnemonics were written by a professional copyeditor and creative writer with a Bachelor of Science degree that we hired on Upwork. As part of the interview, we asked the writer to produce two sample mnemonics to ensure the mnemonics would be high quality, and the annotator was paid a high rate of \$3 per mnemonic (around \$60/hr), which is fair for the participant demographic.

\begin{figure*}
    \centering
    \fbox{\includegraphics[width=\linewidth]{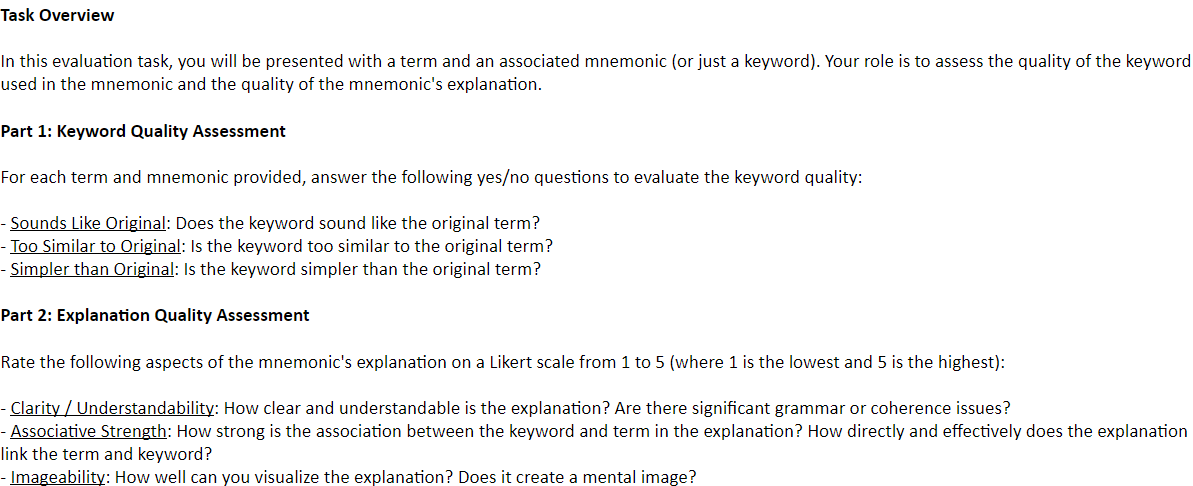}}
    \caption{Instructions to expert educators when rating the keyword and explanation quality of mnemonic devices.}
    \label{fig:annotator}
\end{figure*}

Our annotators who rated the keyword and explanation quality of these mnemonics are both researchers in memory and mnemonic research (one post-doc and one assistant professor). The instructions given to these annotators are shown in Figure~\ref{fig:annotator}. Annotators were paid at a rate of \$50/hr, which is fair for the participant demographic. Our annotators showcase moderate agreement, highlighting the subjective nature of mnemonic generation (Table~\ref{table:qual_agreement}). Numerical tabular versions of the results from Figure~\ref{fig:human} are in Tables~\ref{table:qual1_num} and \ref{table:qual2_num}.

\subsection{Learning Time Distribution} \label{appendix:learning_time}

We provide the distribution of learning time, measured in turns needed to recall the definition of the vocabulary term, for the winning and losing mnemonics in the pair.
As expected, the winning mnemonics have a significantly lower ($p=0.05$) average number of turns needed until the definition is recalled, showing a clear gap in short-term learning efficacy between mnemonics in our pairs. 

\section{Detailed Analysis and Results} \label{appendix:results}

\subsection{Offensive Mnemonics can Aid Learning} \label{appendix:offensive_learning}
\textbf{\textcolor{red}{Warning: This subsection contains an example of an offensive mnemonic.}}

In \cref{section:rlhf}, we describe our rationale for using both expressed and observed preferences;
while observed preferences often reflect our downstream goal, expressed preferences ensure that this goal is achieved in a safe manner.
For example, if we were to optimize mnemonics just on learning, which is our downstream goal, we may produce bizarre or offensive mnemonics, since these mnemonics have been shown to help students learn \cite{wollen1987bizarreness}.
However, expressed preferences are a more reliable method to detect these offensive mnemonics, and is thus likely why the majority of LLM alignment methods for dialogue safety rely on expressed preferences as training data.

To illustrate this point, we present the following mnemonic generated by the initial \model model during the user study, which was flagged as one of two offensive mnemonics: \emph{Obtuse sounds like "abuse". If you abuse someone, they may not understand the situation, just like an obtuse person who is slow to understand}.
As noted by our annotator, this mnemonic: ``\emph{may be insensitive to people who have felt abuse in their lives and feel as if the mnemonic is calling them slow}''.
Through expressed preferences, this mnemonic received a Likert rating of 1 and received 0 votes in the pairwise comparison.
However, with observed preferences, the student studying with this mnemonic learned the term in just one iteration.
Thus, while observed preferences would suggest that this mnemonic is highly effective for learning, the expressed preferences show that this mnemonic may be offensive or harmful to users, motivating our use of all preference labels for enhancing mnemonic generation.

\subsection{Are Bayesian Tie-Breaks Good?} \label{appendix:tiebreak}

We previously found that using multiple preference labels to break ties within singular preference labels, improving LLM output quality with DPO (\cref{subsection:rl_eval}).
To ensure that these ties are better than random tie breaks, we use our GPT-4 classifier to compare the winning and losing mnemonics from our Bayesian labels (i.e. mnemonics with higher and lower effectiveness) where the majority vote in the pairwise setting is \texttt{Tie}.
On the set of ties, GPT-4 states that our ``winning'' mnemonics is better 40\% of the time, tied with the losing mnemonic 31\% of the time, and is worse 29\% of the time.
Thus, even though humans marked these LLM outputs as tied, we were able to draw from other preference labels to identify a winning mnemonic, and GPT-4 also tends to agree that this mnemonic is higher quality.

For context, when students mark a mnemonic as ``winning'' from pairwise comparisons (i.e. non-tie), GPT-4 states this mnemonic is higher-quality than the alternative in 51\% of cases, tied in 22\% of cases, and is lower-quality in 26\% of cases. 

\subsection{DPO Models versus Fine-Tuning} \label{appendix:dpo_sft}

In Table~\ref{table:quant_sft}, we use GPT-4 to judge the mnemonics produced by each of the DPO configurations used in \cref{subsection:rl_eval} versus the \model model just using fine-tuning $p_0(m | v)$. We find that DPO improves the mnemonic quality of each of these models.

\subsection{Mnemonic Examples} \label{appendix:examples}

In Table~\ref{table:mnemonics}, we provide examples of high-quality keyword mnemonic devices generated by our final \model model. We also show some examples of low-quality mnemonics from our model and highlight areas for improvement in Table~\ref{table:mnemonics_bad}.

\begin{table*}[]
\footnotesize
\centering
\begin{tabular}{@{}l|l|c@{}}
\toprule
\multicolumn{1}{c|}{\textbf{Column}} & \multicolumn{1}{c|}{\textbf{Description}} & \multicolumn{1}{c}{\textbf{Num Unique}} \\ \midrule
term & Vocabulary term & 822 \\
mnemonic & Mnemonic for the vocabulary term & 889 \\ \bottomrule
\end{tabular}
\caption{Descriptions of columns in $\mathcal{D}_{train}$.}
\label{table:train}
\end{table*}

\begin{table*}[t]
\centering
\setlength{\tabcolsep}{5pt}
\begin{tabular}{@{}lccc@{}}
\toprule
\textbf{Preference Type} & \textbf{\# Annotations} & \textbf{\# Pairs} & \textbf{Average Annotations / Pair} \\ \midrule
Pairwise ($y_{pair}$) & 1693 & 460 & 3.68 \\
Rating ($y_{rate}$) & 389 & 121 & 3.21 \\
Learning ($y_{learn}$) & 602 & 170 & 3.54 \\ \bottomrule
\end{tabular}
\caption{\label{table:summary} Summary statistics of preference annotations.}
\end{table*}

\begin{table*}[]
\footnotesize
\centering
\begin{tabular}{@{}l|l|c@{}}
\toprule
\multicolumn{1}{c|}{\textbf{Column}} & \multicolumn{1}{c|}{\textbf{Description}} & \multicolumn{1}{c}{\textbf{Num Unique}} \\ \midrule
term & Vocabulary term & 472 \\
mnemonic\_A & Mnemonic A for the vocabulary term & 472 \\
mnemonic\_B & Mnemonic B for the vocabulary term & 472 \\
pairwise\_A\_votes & Number of users who picked Mnemonic A in the pairwise comparison & 11 \\
pairwise\_B\_votes & Number of users who picked Mnemonic B in the pairwise comparison & 9 \\
pairwise\_tie\_votes & Number of users who picked ``tie'' in the pairwise comparison & 5 \\
A\_likert\_ratings & List of Likert ratings from 1-5 denoting the quality of Mnemonic A & 5 \\
B\_likert\_ratings & List of Likert ratings from 1-5 denoting the quality of Mnemonic B & 5 \\
A\_learn\_iterations & List of turns from 1 to $\infty$ the student needed to learn the term with Mnemonic A & 7 \\
B\_learn\_iterations & List of turns from 1 to $\infty$ the student needed to learn the term with Mnemonic B & 8 \\ \bottomrule
\end{tabular}
\caption{Descriptions of columns in $\mathcal{D}_{pref}$.}
\label{table:pref}
\end{table*}

\begin{table*}[]
\centering
\begin{tabular}{@{}cccc@{}}
\toprule
\textbf{Feedback} & \textbf{Metric Used} & \textbf{Preference Agreement} & \textbf{Random Agreement} \\ \midrule
Comparison & Average Entropy & 0.802 & 1.222 \\
Rating & Average Variance & 0.778 & 1.331 \\
Learning & Average Variance & 0.323 & 2.179 \\ \bottomrule
\end{tabular}
\caption{Quantifying annotation noise in  $\mathcal{D}_{pref}$.}
\label{table:noise}
\end{table*}

\begin{table*}[]
\centering
\begin{tabular}{@{}c|c@{}}
\toprule
\textbf{Model Qualitative Comparison}   & \textbf{Cohen's $\kappa$} \\ \midrule
\model vs Transphoner (PS)              & 0.381                     \\
\model vs Transphoner (Simplicity)      & 0.526                     \\ \midrule
\model vs GPT-4 vs Human (PS)           & 0.497                     \\
\model vs GPT-4 vs Human (Simplicity)   & 0.538                     \\ \midrule
\model vs GPT-4 vs Human (Clarity)      & 0.428                     \\
\model vs GPT-4 vs Human (Strength)     & 0.356                     \\
\model vs GPT-4 vs Human (Imageability) & 0.681                     \\ \bottomrule
\end{tabular}
\caption{Agreement of our annotators across all qualitative evaluations (model comparisons and aspects). We measure agreement through Cohen $\kappa$, using quadratic weighting for the Likert scale ratings.}
\label{table:qual_agreement}
\end{table*}

\begin{table*}[]
\centering
\begin{tabular}{@{}ccc@{}}
\multicolumn{1}{l}{}                & \multicolumn{2}{c}{\textit{Keyword Quality}}         \\ \midrule
\multicolumn{1}{c|}{\textbf{Model}} & Phonetic Similarity & \multicolumn{1}{l}{Simplicity} \\ \midrule
\multicolumn{1}{c|}{\model}         & 0.87                & 0.75                           \\
\multicolumn{1}{c|}{Transphoner}    & 0.87                & 0.73                           \\ \bottomrule
\end{tabular}
\caption{Numerical tabular version of qualitative evaluation of \model versus Transphoner from Table~\ref{fig:human}.}
\label{table:qual1_num}
\end{table*}

\begin{table*}[]
\centering
\begin{tabular}{@{}cccccc@{}}
\multicolumn{1}{l}{}                & \multicolumn{2}{c}{\textit{Keyword Quality}}          & \multicolumn{3}{c}{\textit{Explanation Quality}}                                              \\ \midrule
\multicolumn{1}{c|}{\textbf{Model}} & Phonetic Similarity & \multicolumn{1}{l|}{Simplicity} & \multicolumn{1}{l}{Clarity} & \multicolumn{1}{l}{Strength} & \multicolumn{1}{l}{Imageability} \\ \midrule
\multicolumn{1}{c|}{\model}         & 0.86                & \multicolumn{1}{c|}{0.67}       & 3.10                        & 2.86                         & 1.91                             \\
\multicolumn{1}{c|}{GPT-4}          & 0.94                & \multicolumn{1}{c|}{0.58}       & 3.17                        & 2.74                         & 2.52                             \\
\multicolumn{1}{c|}{Human}          & 0.96                & \multicolumn{1}{c|}{0.91}       & 3.64                        & 2.99                         & 3.38                             \\ \bottomrule
\end{tabular}
\caption{Numerical tabular version of qualitative evaluation of \model versus GPT-4 versus Humans from Table~\ref{fig:human}.}
\label{table:qual2_num}
\end{table*}

\begin{table*}[]
\centering
\begin{tabular}{@{}ccc@{}}
\toprule
Percentile & Winning Mnemonic Turns & Losing Mnemonic Turns \\ \midrule
0 & 1.00 & 1.25 \\
25 & 1.00 & 1.60 \\
50 & 1.00 & 2.00 \\
75 & 1.50 & 2.00 \\
100 & 3.00 & 7.00 \\ \midrule
\textbf{Average} & \textbf{1.24} & \textbf{2.13} \\ \bottomrule
\end{tabular}
\caption{Distribution of learning time (Turns) for winning and losing mnemonics.}
\label{table:learning_time}
\end{table*}

\begin{table*}[]
\centering
\begin{tabular}{@{}cccc@{}}
\toprule
\textbf{Model A/B Pair} & \textbf{A Wins} & \textbf{Tie} & \textbf{B Wins} \\ \midrule
$p_{pair}(m | v)$ (Table~\ref{table:quant}, top), \; $p_{0}(m | v)$ & \textbf{0.26} & 0.54 & 0.20 \\
$p_{bayes}(m | v)$ (Table~\ref{table:quant}, top), \; $p_{0}(m | v)$ & \textbf{0.22} & 0.58 & 0.20 \\
$p_{pair}(m | v)$ (Table~\ref{table:quant}, bottom), \; $p_{0}(m | v)$ & \textbf{0.24} & 0.58 & 0.18 \\
$p_{dpo}(m | v)$ (Table~\ref{table:quant}, bottom), \; $p_{0}(m | v)$ & \textbf{0.29} & 0.53 & 0.18 \\
\bottomrule
\end{tabular}
\caption{GPT judgement of DPO models versus fine-tuning. The better model is in \textbf{bold}.}
\label{table:quant_sft}
\end{table*}

\begin{table*}[ht]
\small
\centering
\begin{tabularx}{\linewidth}{c | X }
\toprule
\textbf{Term} & \multicolumn{1}{c}{\textbf{Generated Keyword Mnemonic}} \\ \midrule

Lionized  & \specialcellleft{Lionized sounds like "lion-eyes," envisioning a lion being admired for its eyes.\\Lionized means to be admired or treated like a celebrity.} \\ \midrule

Empiricism  & \specialcellleft{Empiricism sounds like "emperor";\\an emperor may rely on experience rather than theory to rule.\\Empiricism is based on experience rather than theory.} \\ \midrule

Dwindling  & \specialcellleft{Dwindling sounds like "wind" + "lingering";\\when the wind lingers, it gradually decreases in strength.\\This relates to dwindling, which means decreasing in size, amount, or strength.} \\ \midrule

Glowered  & \specialcellleft{Glowered sounds like "glow" + "red".\\Imagine a red glowing face, which could represent anger or a glowered expression.} \\ \midrule

Static  & \specialcellleft{Static sounds like "stick", which is stationary or unmoving.\\Static electricity is also stationary, as it does not move. \\Both relate to the concept of being stationary or unmoving.} \\ \midrule

Sanctimony  & \specialcellleft{Sanctimony sounds like "saint" + "money";\\saints are supposed to be holy, but some may be hypocritical and greedy for money.\\ This hypocrisy and false piety is sanctimony.} \\ \midrule

Goosebumps  & \specialcellleft{Goosebumps sounds like "ghost bumps," which can happen when you see a ghost.\\Goosebumps are bumps on the skin caused by fear or cold.} \\ \midrule

Resurgence  & \specialcellleft{Resurgence sounds like "re-surgery";\\after a surgery, one may experience a resurgence of pain.\\Resurgence refers to a renewed increase or rise..} \\ \midrule

Moot  & \specialcellleft{Moot sounds like "mute";\\ when something is mute, it's debatable whether it's on or off.\\Similarly, a moot point is debatable.} \\ \midrule

Subordinate & \specialcellleft{Subordinate sounds like "sub-order", implying a lower rank or position.\\ Subordinates are lower in rank or position, and are subject to the authority of others.} \\ \midrule

Upbraided & \specialcellleft{Upbraided sounds like "up" + "braid".\\Imagine a teacher scolding a student for not braiding her hair properly.\\This scolding or reprimanding is upbraiding.} \\ \midrule

Despoiled & \specialcellleft{Despoiled sounds like "despite spoiled."\\Imagine someone spoiling your plans despite your efforts, leaving you feeling despoiled.} \\ \midrule

Caterwauling & \specialcellleft{Caterwauling sounds like "cat + war + yelling," imagining a cat yelling during war.\\Caterwauling refers to loud, unpleasant noise.} \\ \midrule

Escalate & \specialcellleft{Escalate sounds like "escalator," which goes up, representing an increase or rise.\\Escalate means to increase or rise in intensity or severity.} \\ \midrule

Compromise & \specialcellleft{Compromise sounds like "com + promise".\\When you make a promise, you may have to compromise on something to keep it.\\For example, you may have to compromise on your sleep to keep a promise\\to meet someone early in the morning.} \\ \midrule

Deftness & \specialcellleft{Deftness sounds like "leftness," which can be associated with left-handed people\\who are often known for their dexterity or deftness.\\This can help remember the meaning of deftness as skillful or adroit.} \\ \midrule

Haptic & \specialcellleft{Haptic sounds like "happy tic", which can be associated with a happy tickle, relating to touch.\\Haptic refers to the sense of touch.} \\ \midrule

Parenthetical & \specialcellleft{Parenthetical sounds like "parenthesis," which is used to add extra information in a sentence.\\Parenthetical means providing additional information in a sentence.} \\ \midrule

Fillip & \specialcellleft{Fillip sounds like "fill up";\\when you fill up a glass, it gives a sudden push to the liquid, similar to a fillip.\\A fillip is a sudden push or stimulus.} \\ \midrule

Kowtow  & \specialcellleft{Kowtow sounds like "cow tow"---imagine a cow towing a car, symbolizing submission or obeisance.\\Kowtow means to show respect or submission by bowing or kneeling.} \\ \bottomrule

\end{tabularx}
\caption{\label{table:mnemonics} Examples of 20 high-quality keyword mnemonics generated by our final model $p_{dpo}(m | v)$.}
\end{table*}

\begin{table*}[ht]
\small
\centering
\begin{tabularx}{\linewidth}{c | X }
\toprule
\textbf{Term} & \multicolumn{1}{c}{\textbf{Generated Keyword Mnemonic}} \\ \midrule

Affiliated & \specialcellleft{Affiliated sounds like \textbf{\textcolor{red}{"affiliation,"}}\\which is a connection or association with a group or organization.\\Being affiliated means being connected or associated with something.} \\ \midrule

Evinced & \specialcellleft{Evinced sounds like "evidence";\\evidence is proof, which is what evinced means.\\\textbf{\textcolor{red}{Evidence is also a synonym of evinced}}} \\ \midrule

Complicit & \specialcellleft{Complicit sounds like \textbf{\textcolor{red}{"compliant" + "licit"}} (licit means legal).\\If someone is compliant with legal actions, they are involved in them.\\Complicit means involved in a wrongdoing.} \\ \midrule

Quintessential & \specialcellleft{Quintessential sounds like "quint" (five) + "essential". \\ \textbf{\textcolor{red}{Five essentials are required to make a perfect dish}}, representing the quintessential.} \\ \midrule

Peons  & \specialcellleft{Peons sounds like \textbf{\textcolor{red}{"pions" (pions are subatomic particles)}}.\\Pions are subatomic particles, so peons are subordinate workers.} \\ \bottomrule

\end{tabularx}
\caption{\label{table:mnemonics_bad} Examples of low-quality keyword mnemonics generated by our final model $p_{dpo}(m | v)$. Prominent issues and areas for improvement are highlighted in \textbf{\textcolor{red}{red}}.}
\end{table*}

\end{document}